\documentclass[letterpaper]{article} %
\usepackage{aaai2026}  %
\usepackage{times}  %
\usepackage{helvet}  %
\usepackage{courier}  %
\usepackage[hyphens]{url}  %
\usepackage{graphicx} %
\usepackage{natbib}  %
\usepackage{caption} %
\usepackage{algorithm}
\usepackage{algorithmic}

\usepackage{newfloat}
\usepackage{listings}
\DeclareCaptionStyle{ruled}{labelfont=normalfont,labelsep=colon,strut=off} %
\floatstyle{ruled}
\newfloat{listing}{tb}{lst}{}
\floatname{listing}{Listing}
\usepackage{amsmath,amssymb}
\usepackage[binary-units=true]{siunitx}
\usepackage{physics}
\usepackage{subcaption}
\usepackage{booktabs}
\usepackage{multirow}
\usepackage{colortbl}
\usepackage{xcolor}
\usepackage{threeparttable}

\newtheorem{definition}{Definition}

\long\def\invis#1{}
\newcommand{\vworld}{\mathbf{V}_{\mathrm{world}}}
\newcommand{\vcurrent}{\mathbf{V}_{\mathrm{c}}}
\newcommand{\vthrust}{\mathbf{V}_{\mathrm{thrust}}}
\newcommand{\turnmaxspd}{\omega_{max}}
\newcommand{\currentfield}{\mathbf{C}}
\newcommand{\navigablearea}{\mathcal{N}}
\newcommand{\nogozone}{\mathcal{L}}
\newcommand{\priormap}{\mathcal{M}}
\newcommand{\homotopicchannel}{\mathcal{H}}
\newcommand{\softnogozone}{\mathcal{V}}
\newcommand{\rrt}{RRT}
\newcommand{\rrtstar}{RRT$^\star$}
\newcommand{\rrtdubins}{RRT-D}
\newcommand{\rrtstardubins}{RRT$^\star$-D}
\newcommand{\prm}{PRM}
\newcommand{\prmdubins}{PRM-D}
\newcommand{\gridastar}{grid A$^\star$}
\title{RENEW: Risk- and Energy-Aware Navigation in Dynamic Waterways}
\author {
    Mingi Jeong\textsuperscript{\rm 1,\rm 2}, Alberto Quattrini Li\textsuperscript{\rm 2}
}
\affiliations {
    \textsuperscript{\rm 1}Department of Aerospace and Ocean Engineering, Virginia Tech\\
    \textsuperscript{\rm 2}Department of Computer Science, Dartmouth College\\
    mingijeong@vt.edu, alberto.quattrini.li@dartmouth.edu
}

\usepackage{bibentry}

\begin{document}

\maketitle

\begin{abstract}
We present RENEW, a global path planner for Autonomous Surface Vehicle (ASV) in dynamic environments with external disturbances (e.g., water currents). 
RENEW introduces a unified risk- and energy-aware strategy that ensures safety by dynamically identifying non-navigable regions and enforcing adaptive safety constraints.
Inspired by maritime contingency planning, it employs a best-effort strategy to maintain control under adverse conditions. The hierarchical architecture combines high-level constrained triangulation for topological diversity with low-level trajectory optimization within safe corridors.
Validated with real-world ocean data, RENEW is the first framework to jointly address adaptive non-navigability and topological path diversity for robust maritime navigation.
\end{abstract}

\begin{links}
    \link{Code}{https://github.com/dartmouthrobotics/RENEW.git}
\end{links}
\section{Introduction}

This paper presents a novel global path planner for Autonomous Surface Vehicle (ASV), addressing external disturbances like water currents, which significantly impact \textbf{navigational risk} and \textbf{energy cost}.
For instance, in the Malacca and Singapore Straits—among the world's busiest waterways—dynamic surface currents reshape navigable areas and dictate optimal paths.
To mitigate risk, vessels must avoid adverse currents while ensuring turning maneuverability allows for reliable contingency maneuvers.

\begin{figure}[t]
    \centering
    \includegraphics[width=0.9\columnwidth]{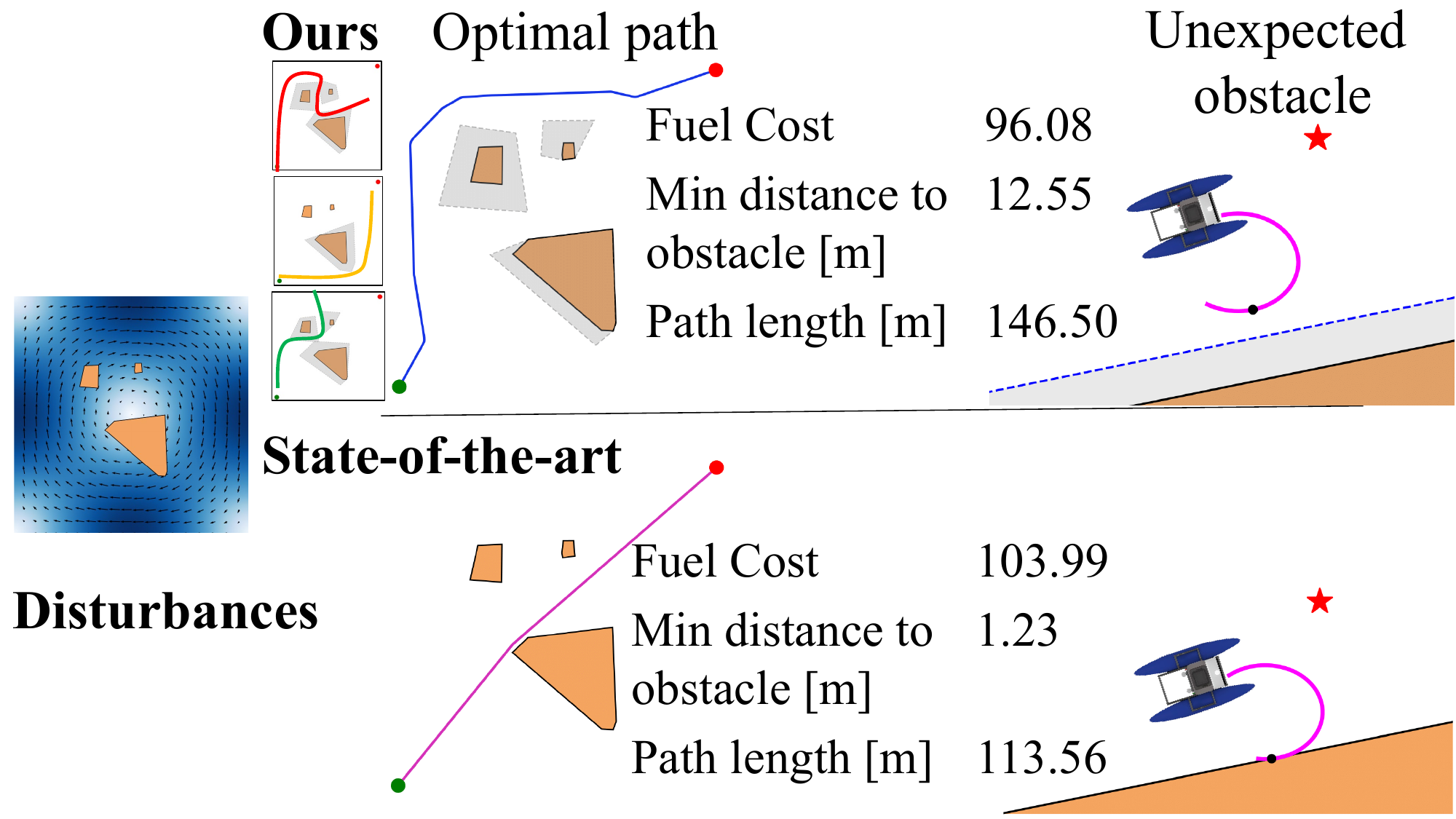}
    \caption{Path planning under external disturbances. %
    (\textbf{\textit{top}}) Our method selects energy-efficient paths across multiple homotopy classes, using current-based adaptive padding (gray) around obstacles (brown), to ensure feasible contingency maneuvers.
    (\textbf{\textit{bottom}}) The baseline prioritizes distance over energy/homotopy. Without adaptive padding, its paths lack safety margins for contingency maneuvers.
    }
    \label{fig:beauty}
\end{figure}

Global path planning under dynamic disturbances remains a challenge, particularly in \textbf{identifying navigable areas} and performing \textbf{safety analysis} to prevent grounding. To ensure ASVs respect state constraints, we propose a risk- and energy-aware framework that guarantees a feasible \textbf{contingency maneuver}—defined as a \textbf{best effort under the worst case}—to avoid Inevitable Collision States (ICS)~\cite{blackmore2011chance, johnson2021chance}. Our approach keeps trajectories within navigable regions despite adverse conditions and bounded uncertainty.

Prior work has addressed path planning under disturbances like ocean currents~\cite{hollinger-risk-aware-2013,Kularatne-RSS-16,stranding-cdc-2023}, yet optimizing across multiple distinct paths while considering dynamic, restricted navigable areas remains an open challenge. A gap exists in accounting for non-holonomic constraints and irregular maneuvering (e.g., current-dependent turning radii) within these changing environments.

Given an environment map and vector field (e.g., ocean currents), our planner incorporates: (1) \textbf{risk-aware safety} via adaptive padding to avoid \textbf{non-navigable areas}; (2) a \textbf{best-effort strategy} for worst-case disturbances; and (3) a hierarchical architecture where a high-level planner identifies \textbf{topologically distinct} (homotopic) paths via constrained triangulation, and a low-level planner optimizes for energy and kinematics (see Fig.~\ref{fig:beauty}).

In summary, our key contributions are:
\begin{itemize}
    \item A safety framework using best-effort turning under uncertainty, accounting for irregular ASV maneuverability and current-dependent no-go areas;
    \item An efficient hierarchical planner that uses constrained triangulation to find topologically distinct paths, optimizing for energy efficiency and kinematic feasibility;
    \item Extensive validation via ablation studies and simulations in both custom realistic and real-world current scenarios.
\end{itemize}
This is the first risk-aware global planner to unify dynamic external forces with topological guarantees for improved safety and efficiency.
\section{Related Works}

Plethora of methods for \textbf{global path planning} appeared in the robotics literature, including efficient representations of the environment through sampling-based approaches, such as PRM \cite{prm-Kavraki-1996} and RRT~\cite{lavalle2001randomized}, and corresponding search-based methods that can find the optimal path, e.g., A*~\cite{hart1968formal}.  For a general overview on path planning and specifically for ASVs, please refer to~\cite{kavraki2016motion} and \cite{vagale2021path, vagale2021path2}, respectively. Here we specifically discuss the relatively more recent global path planners that consider to some extent uncertainty and external disturbances, as well as related work that inspires our work, particularly tube-based and homotopic approaches.

\textbf{Uncertainty} in path planning arises from two sources: (1) imperfect prior knowledge of the environment and (2) deviations during execution due to motion/sensing noise or controller limitations. Some approaches model this uncertainty through collision probability bounds on roadmaps~\cite{guibas2010bounded}, enabling chance-constrained optimization that bounds \textit{collision risk}~\cite{blackmore2011chance, johnson2021chance}. For non-holonomic vehicles, the notion of \textbf{non-navigable area} captures regions where the vehicle may reach an unsafe state~(ICS) despite its best effort \cite{viability-Fezari-2019, viability-vessel-2015}. The boundary's shape can depend on uncertain disturbances, making its identification an open challenge. 
We adaptively identify non-navigable areas based on the ASV’s encounter direction within the external disturbance field, taking into account the vehicle’s non-holonomic constraints and probabilistically reasoning about worst-case scenarios---mirroring the maritime concept of an \textit{abort position}, where fallback options for the safety are ensured under adverse conditions.

\textbf{External disturbances} have been mostly considered in the context of control and trajectory tracking, with methods based on 
feedback tracking controller and
a reconfigurable disturbance compensation mechanism \cite{liu2018adaptive}, Generalized Reduced Gradient \cite{rudd2017generalized}, and reinforcement learning \cite{faust2015preference,blekas2018rl}. 
In global path planning, some methods set an additional fixed safety distance that accounts for currents \cite{singh2018constrained}. 
\citet{aine2016integrating} integrates A* with a controller for feasibility checking (including disturbances) of connecting cells along the path.
Trajectory generation using Hamilton-Jacobi reachability analysis to find error propagation due to disturbance can provide safety guarantees within a prediction horizon (e.g., for drones \cite{herbert2017fastrack,seo2019robust}).
External disturbances are also considered for optimizing the path energy efficiency. 
\citet{jones2017planning} proposed a stochastic trajectory optimization approach for underwater vehicles.
An approach based on graphs with a flow model and cost function that includes energy finds paths for ASVs and AUVs so that they can leverage the dynamics of the surrounding flow \cite{kularatne2018going}.
A two-stage planner composed of A* and a solver for an optimal continuous problem uses an energy cost function that considers wind to find energy-efficient paths \cite{bitar2020two}.
We solve instead a global path planning problem, considering \textbf{both} feasibility, which is dynamically determined based on external disturbances, and energy efficiency.

\textbf{Path boundaries} were proposed for safe path planning for swarm of robots in cluttered environments within a single homotopic option based on the tube~\cite{mao2024optimal} or path set~\cite{Huang-homotopy-2024}. 
Similarly, robust motion planners leveraged the idea of funnels, which provides safety guarantees \cite{singh2017robust,majumdar2017funnel}.
The concept of such boundaries inspires our method to account for the uncertainty and external disturbances described above, integrated in global path planning. 
Previous approaches utilized \textbf{homotopy classes} \cite{bhattacharya2012topological} to facilitate multirobot exploration~\cite{kularatne2018going,Huang-homotopy-2024}. We explicitly keep track of such homotopy classes during planning phase for maintaining alternatives the ASV can commit/replan, allowing for adaptive current padding per homotopy, thus increasing ASV safety.

\section{Problem Formulation}
The problem is finding an optimal path $\Gamma^*$ that is both \textbf{safe} and \textbf{energy-efficient}:
\begin{equation}
    \Gamma^* = \underset{\Gamma}{\arg\min} \quad J(\Gamma)
\end{equation}
where $\Gamma$ is a path in the spatial environment $\mathbb{W} \subset \mathbb{R}^2$ and $J(\cdot)$ is the objective function representing energy cost, influenced by fuel consumption under the disturbance field. 

The disturbance field is denoted as $\currentfield(t, \mathbf{x})$ which acts on the ASV at position $\mathbf{x} \in \mathbb{R}^2$ and time $t$. The navigable area at time $t$, denoted $\navigablearea(t)$, is the subset of $\mathbb{W}$ where the ASV can safely operate under the effect of $\currentfield$. 

We consider a marine vehicle with a non-holonomic kinematic model and constant thrust, i.e., constant effort along the forward direction, while $\omega$ action for the directional change can be taken within $[-\turnmaxspd, +\turnmaxspd]$. This maneuvering behavior is typical for vessels operating at a fixed RPM. The resulting effective velocity in the world frame is:
\begin{equation} \label{eq:velocity}
    \vworld(\mathbf{x},t) = \vcurrent(\mathbf{x},t) + \vthrust(\mathbf{x},t)
\end{equation}
where $\vworld$ is the net velocity relative to the ground, $\vthrust$ is the velocity relative to the water (i.e., propulsion), and $\vcurrent$ is the current-induced velocity at $\mathbf{x}$. 

Unlike prior works (e.g., \cite{Kularatne-RSS-16, stranding-cdc-2023}) that assume underactuated dynamics, we assume $||\vthrust(\mathbf{x},t)|| > ||\vcurrent(\mathbf{x},t) ||$: this means the robot can overcome adverse currents, while still being constrained by the non-holonomic dynamics observed in practical scenarios. This formulation captures real-world navigational concerns---such as minimizing fuel consumption in the presence of misaligned currents---while ensuring the bounded safety, even in narrow or risk-prone passages.

We assume the map $\priormap$ and the 2D vector field $\currentfield$ are available a priori (e.g., from forecasts) and remain unchanged during our short navigation time spans \cite{Fossen_2021}. %
We demonstrate that different currents in the same region at different times can result in varying optimal paths. 
\begin{figure}[t!]
    \centering
    \includegraphics[width=0.97\columnwidth]{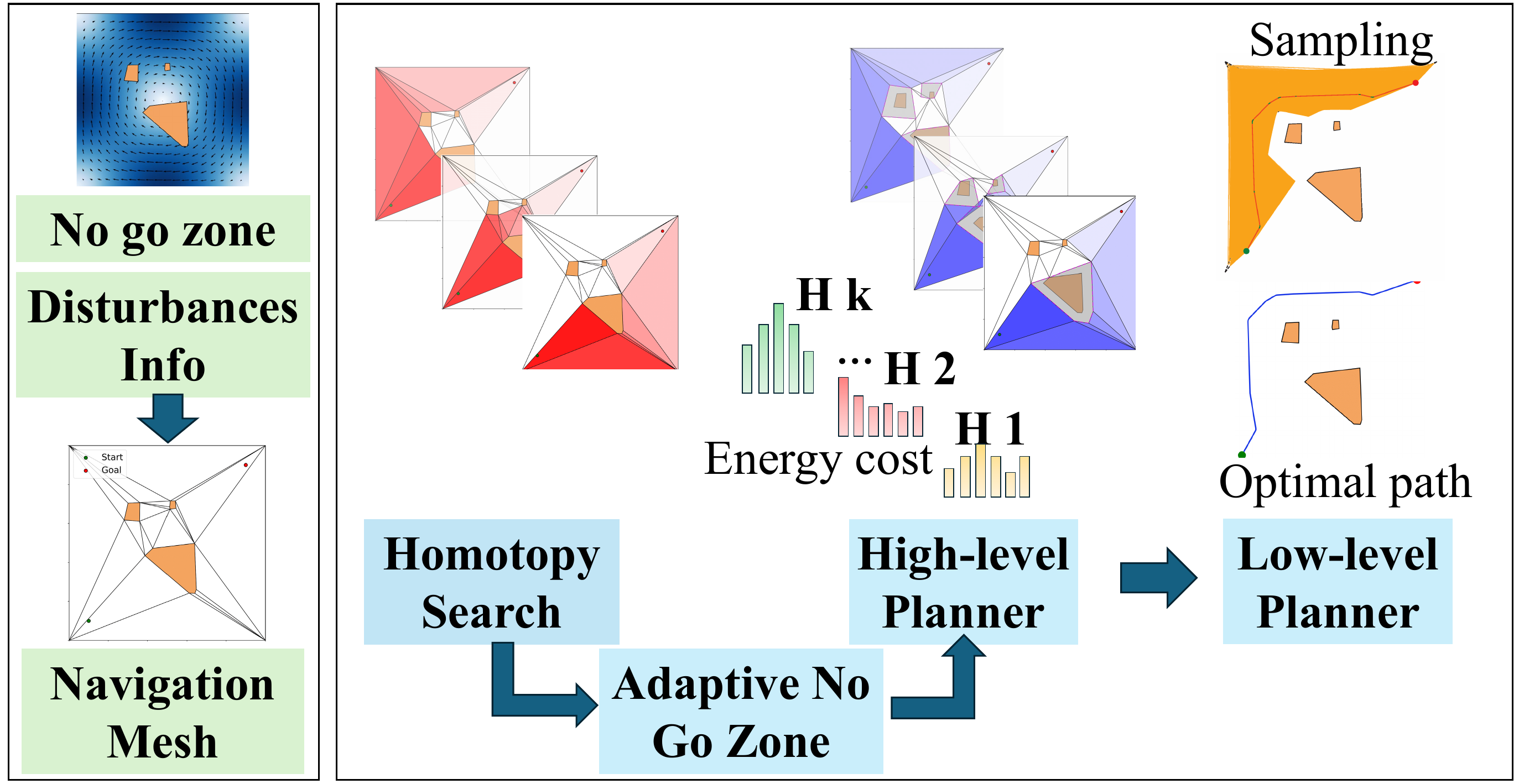}
    \caption{System Architecture.}
    \label{fig:system-architecture}
\end{figure}
\section{Main Approaches} 
RENEW planner's overall architecture is shown in Figure~\ref{fig:system-architecture}.
\subsection{No Go Zone}  We define an unsafe state $\nogozone \subset \mathbb{R}^2$ as \textit{hard} No Go Zone where the ego vehicle $R$ must \textit{never} enter, e.g., shorelines, buoys, and islands. Such $\nogozone$ consists of collections of obstacles, represented by polygonal shapes. $\softnogozone(\tau)$ (\textit{soft} No Go Zone) denotes a set of positions where $R$ may enter $\nogozone$ despite its best effort $\tau$ (i.e., hard-over maneuver) due to, e.g., disturbances.  $\softnogozone(\tau)^c$ is a set of positions where $R$ can avoid $\nogozone$ even if there exist disturbances, to prevent ICS. 

We first construct a set of \textit{navigation meshes} within the complement of the no go zone, denoted as $\nogozone^c$, using convex polygons. To ensure that obstacle boundaries are preserved during meshing, we employ \textit{Constrained Delaunay Triangulation} (CDT)~\cite{Shewchuk-triangle-1996}. %
This constraint-based triangulation partitions the environment into distinct subregions: $\nogozone$ and $\nogozone^c$. We then identify $\softnogozone(\tau)$ by adaptive padding proposed in the next sections.
The resulting navigation meshes provide a structured representation of the robot’s $\navigablearea=\softnogozone(\tau)^c$, facilitating efficient planning. %
Our proposed method varies $\softnogozone(\tau)$ depending on homotopic classes.

\begin{figure}[b]
    \centering
    \begin{minipage}[t]{0.45\columnwidth}
        \begin{subfigure}[b]{\textwidth}
           \includegraphics[width=\textwidth]{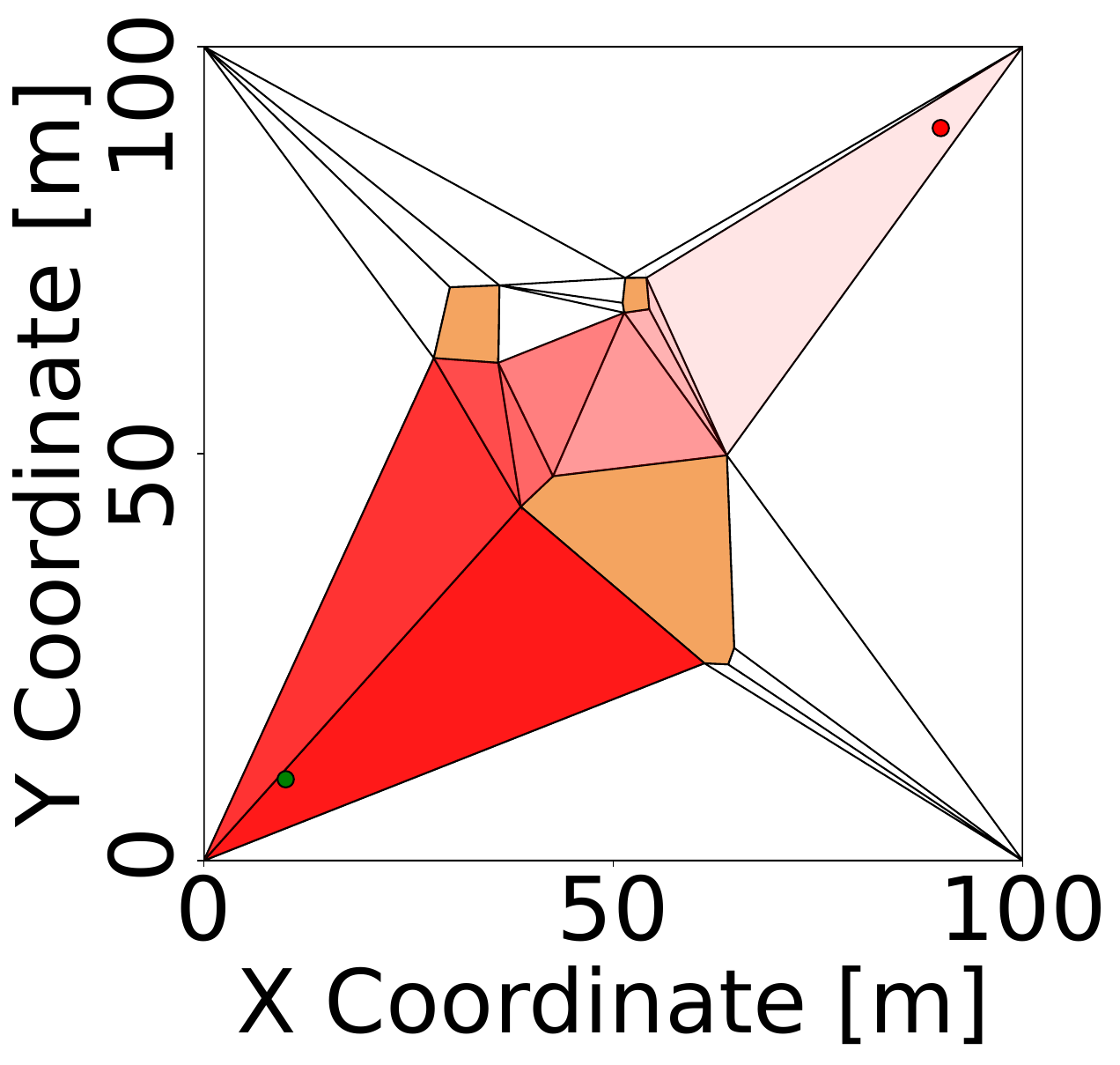}
         \end{subfigure}
    \end{minipage}
    \caption{CDF-based navigation mesh and a single CDT homotopic channel. The color is lighter from the start (10,10) to the goal (90,90). Hard No Go Zone $\nogozone$ is brown.}
    \label{fig:mesh-channel}
\end{figure}

\subsection{Homotopic Channel}
One of our key ideas is to construct a set of homotopic channels $\homotopicchannel$ consisting of multiple homotopy classes found over CDT in $\nogozone^c$ (Figure~\ref{fig:mesh-channel}). Intuitively, one homotopic channel in $\homotopicchannel$ represents a bound that $R$ can follow. Multiple paths belong to a corresponding homotopy class %
if they meet the criteria, i.e., collision avoidance and kinematic feasibility. 

To find such channels, we build roadmap by using dual property of CDT. Specifically, the CDT becomes a node and the edge shared by the neighboring triangles becomes a connection between nodes. Intuitively, from a start point to a goal point, we find a series of triangles, i.e., channel. We use the Depth First Search algorithm that can give one distinct channel, and continue to find topologically distinct channels $\homotopicchannel$ such that $|\homotopicchannel|=k$, where $k$ is the maximum homotopy classes to be found by user input. Note that $k \leq 2^n$ where $n$ is the number of total obstacles in the environment.

\begin{definition} \label{def:channel} (Channel and Homotopic Channel) A series of triangles is defined as a channel that the robot can follow. For each homotopy option, there exists only one set of triangles, i.e., channel.
\end{definition}

In this section, for simplicity, we present an example with obstacles represented as convex polygons. However, as validated in our experiments, the definitions and conditions also hold with concave obstacles by using a crossing-sequence-based method to identify homotopy classes~\cite{Tovar_Cohen_LaValle_2009, Soonkyum_Kim_Likhachev_2015}. %

\begin{figure}[t]
    ~
    \begin{minipage}[t]{0.25\columnwidth}
        \begin{subfigure}[b]{\textwidth}
           \includegraphics[,width=\textwidth]{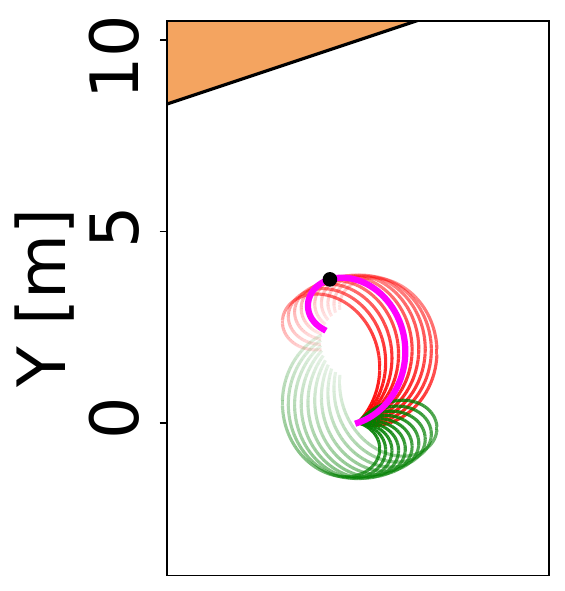}
         \end{subfigure}
    \end{minipage}
    ~
    \begin{minipage}[t]{0.25\columnwidth}
        \begin{subfigure}[b]{\textwidth}
           \includegraphics[,width=\textwidth]{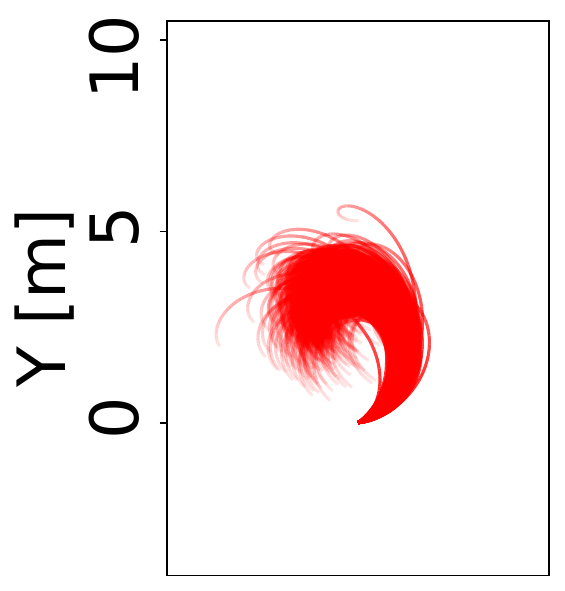}
         \end{subfigure}
    \end{minipage}
    ~
    \begin{minipage}[t]{0.25\columnwidth}
        \begin{subfigure}[b]{\textwidth}
           \includegraphics[width=\textwidth]{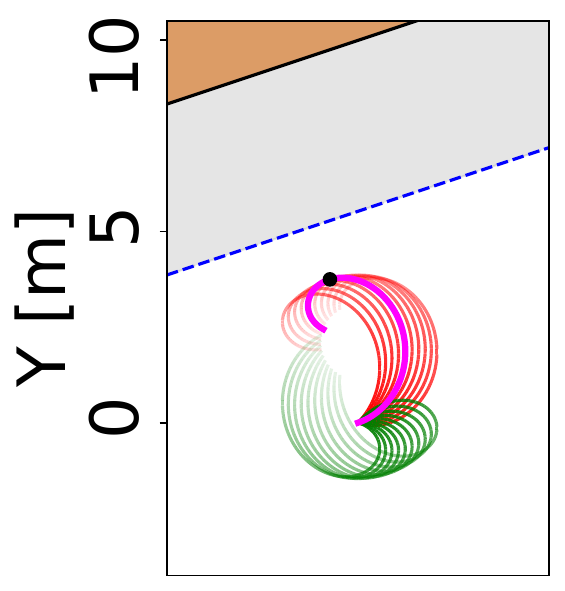}
         \end{subfigure}
    \end{minipage}
    \newline
    \centering
    ~
    \begin{minipage}[t]{0.25\columnwidth}
        \begin{subfigure}[b]{\textwidth}
           \includegraphics[,width=\textwidth]{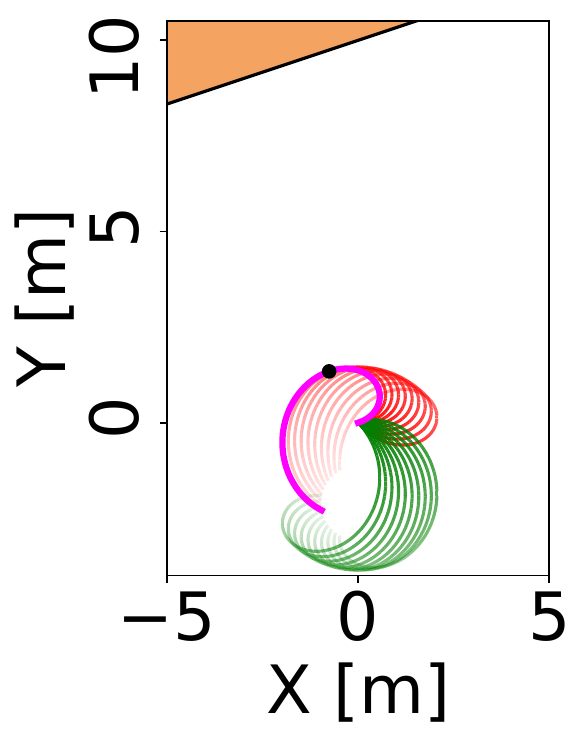}
         \end{subfigure}
    \end{minipage}
    ~
    \begin{minipage}[t]{0.25\columnwidth}
        \begin{subfigure}[b]{\textwidth}
           \includegraphics[,width=\textwidth]{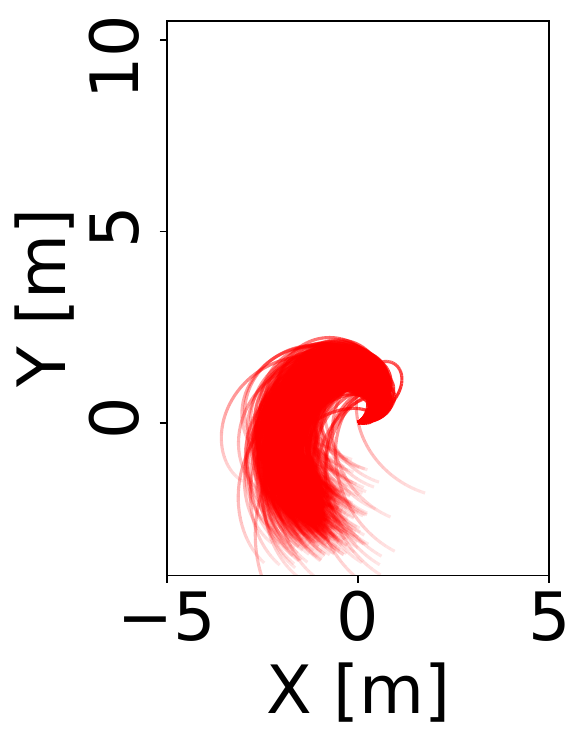}
         \end{subfigure}
    \end{minipage}
    ~
    \begin{minipage}[t]{0.25\columnwidth}
        \begin{subfigure}[b]{\textwidth}
           \includegraphics[width=\textwidth]{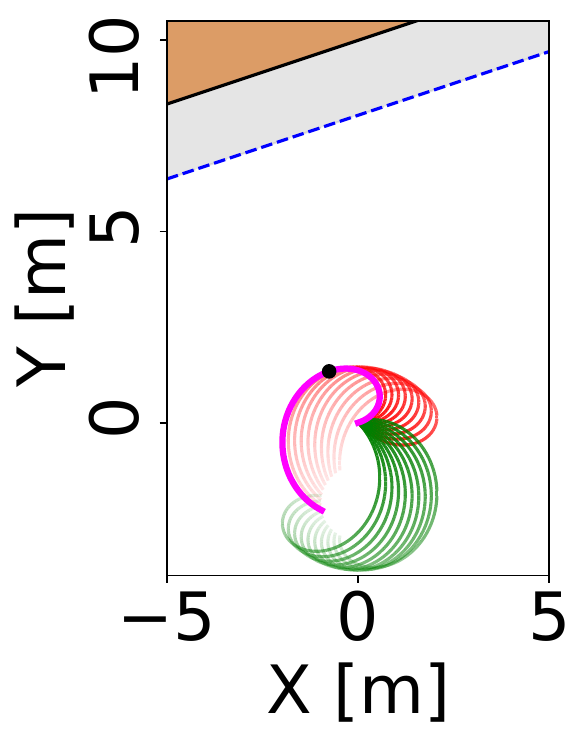}
         \end{subfigure}
    \end{minipage}
    \caption{Irregular turning circle behaviors under the external disturbances with their noises: (\textbf{\textit{top}}) northward current; (\textbf{\textit{bottom}}) southward current; (\textbf{\textit{left}}) turning circles from the robot in the example at (0,0) with heading samples $[\phi-\Delta\phi, \phi+\Delta\phi]$. The closest turning circle (magenta) and its closest point (black dot) to the constrained edge are marked; (\textbf{\textit{mid}}) turning circle behaviors under disturbance noises; (\textbf{\textit{right}}) offset padding $\softnogozone(\tau)$ for the constrained edge (gray), to ensure safety within the probabilistic bound.}
    \label{fig:turning-circle}
\end{figure}

\begin{figure}[t]
    \centering
    \begin{minipage}[t]{0.40\columnwidth}
        \begin{subfigure}[b]{\textwidth} \includegraphics[width=\textwidth]{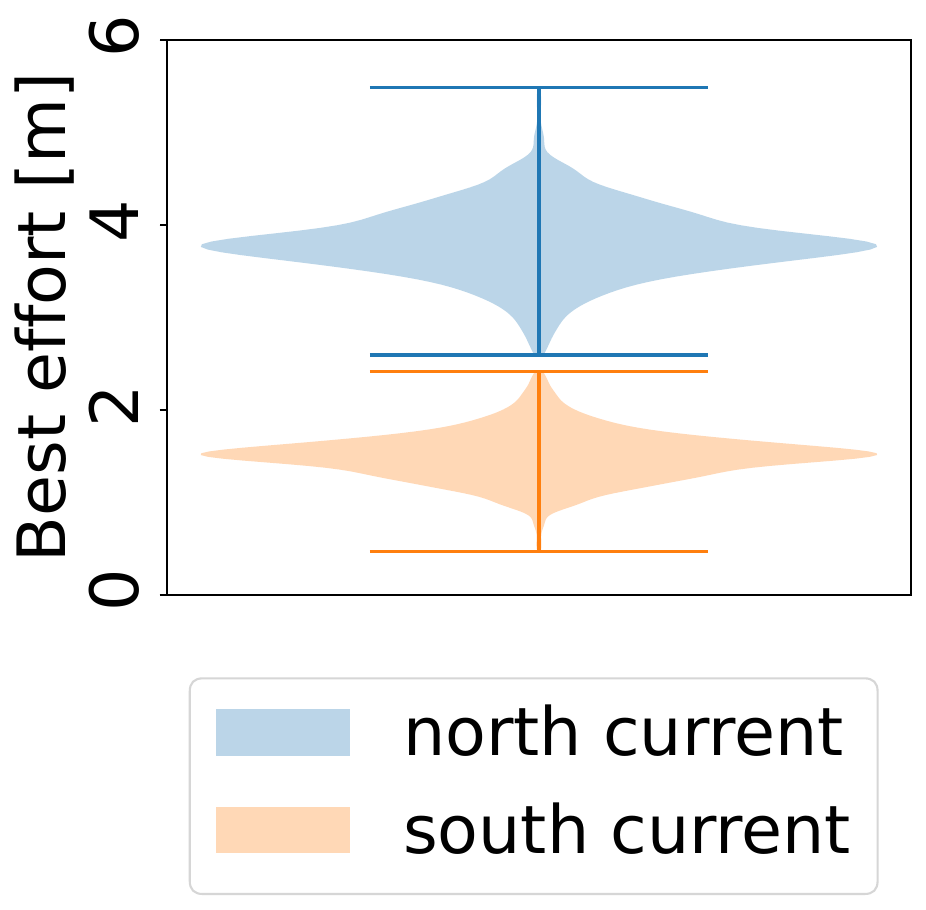}
         \end{subfigure}
    \end{minipage}
    ~
    \begin{minipage}[t]{0.46\columnwidth}
        \begin{subfigure}[b]{\textwidth}
           \includegraphics[width=\textwidth]{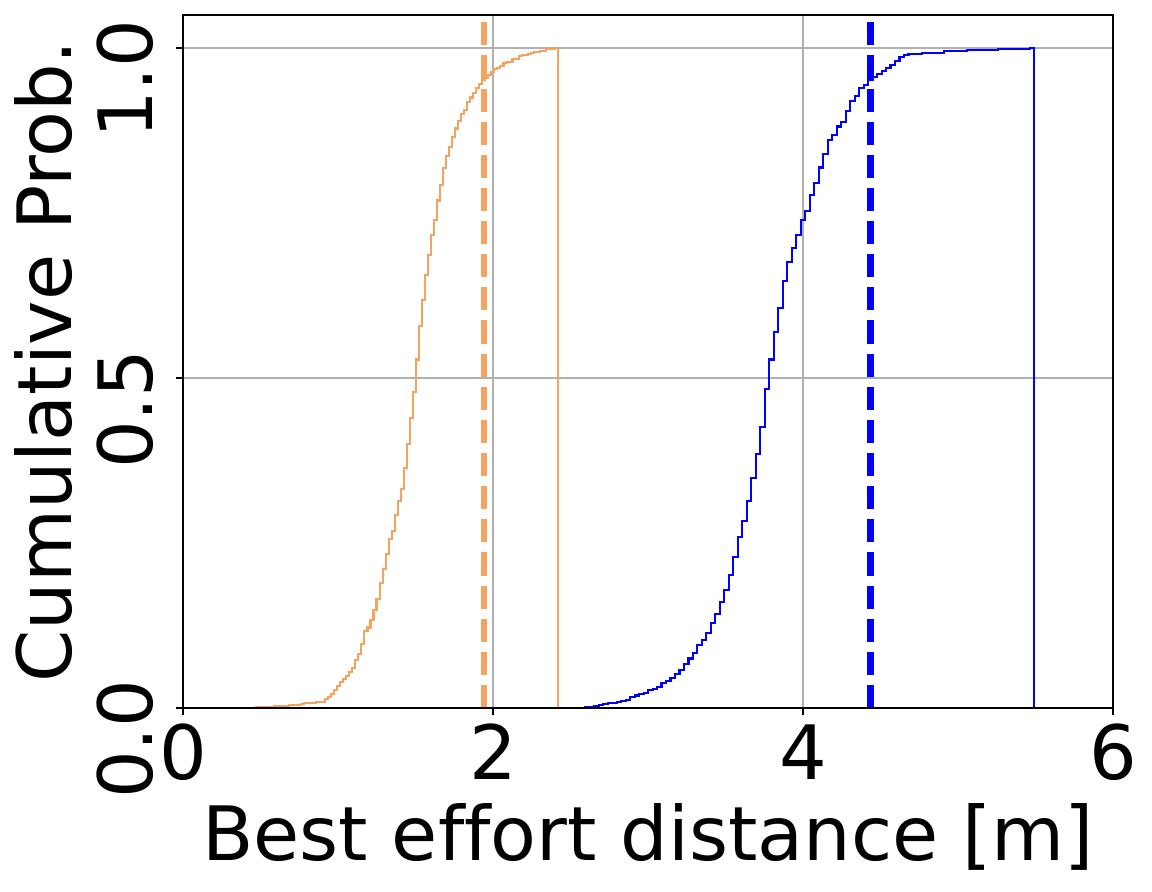}
         \end{subfigure}
    \end{minipage}
    \caption{Probabilistic distribution and bounds of the best-effort maneuvers. (\textbf{\textit{left}}) spatial extent of extreme points induced by best-effort actions; and (\textbf{\textit{right}}) cumulative distribution of best-effort distances, along with bounded values based on the 95th percentile threshold ($\sigma=0.95$ for \SI{4.47}{\meter}: northward current, \SI{1.96}{\meter}: southward current).}
    \label{fig:cdf-best-effort}
\end{figure}
\subsection{Adaptive No Go Zone}

\textbf{Irregular turning:} We consider a turning motion with constant forward speed and a non-zero turning rate $\omega$. Based on Equation~(\ref{eq:velocity}), the system dynamics are given by:
\[
    \dot{p}_x = v_{\text{thrust}} \cos(\theta) + c_{x}, \quad
    \dot{p}_y = v_{\text{thrust}} \sin(\theta) + c_{y}, \quad
    \dot{\theta} = \omega
\]
with $v_{\text{thrust}}$ the thrust-induced speed in the body frame, and $c_x$, $c_y$ the external disturbances %
in the global frame.

We define the best-effort control input $\tau$ as a hard-over turn with angular velocity either $-\turnmaxspd$ or $\turnmaxspd$, to prevent entering the no go zone $\nogozone$. In alignment with naval architecture concepts—specifically \textit{advance} and \textit{tactical diameter}—it is essential to identify the extreme points reached during such turns to ensure the robot remains outside of $\nogozone$.

Without disturbances, the robot traces a regular turning circle with radius $r = v_{\text{thrust}} / \omega$. In the presence of disturbances (i.e., $|c_x| > 0$ or $|c_y| > 0$), the turning shape becomes irregular (Figure~\ref{fig:turning-circle}). In other words, the disturbances affect the turning circle, thus the soft no go zone $\softnogozone(\tau)$.

\noindent\textbf{Adaptive padding:} To identify such varying $\softnogozone(\tau)$, we propose a sampling-based approach to estimate the extreme points under worst-case conditions—when the robot comes closest to a constrained edge of a triangulated obstacle. Specifically, within a triangle, we sample hard-over turns $\tau \in \{-\omega, +\omega\}$ based on: (1) the midline travel direction $\phi$; (2) the cross-track offset $\Delta\phi$, (3) and the average direction and magnitude of  disturbances. Among all sampled turns, we select the path that approaches the constrained edge most closely (i.e., the worst case)--see Figure~\ref{fig:turning-circle} (left).

\begin{figure}[h]
    \centering
    \begin{minipage}[t]{0.45\columnwidth}
        \begin{subfigure}[b]{\textwidth}
           \includegraphics[width=\textwidth]{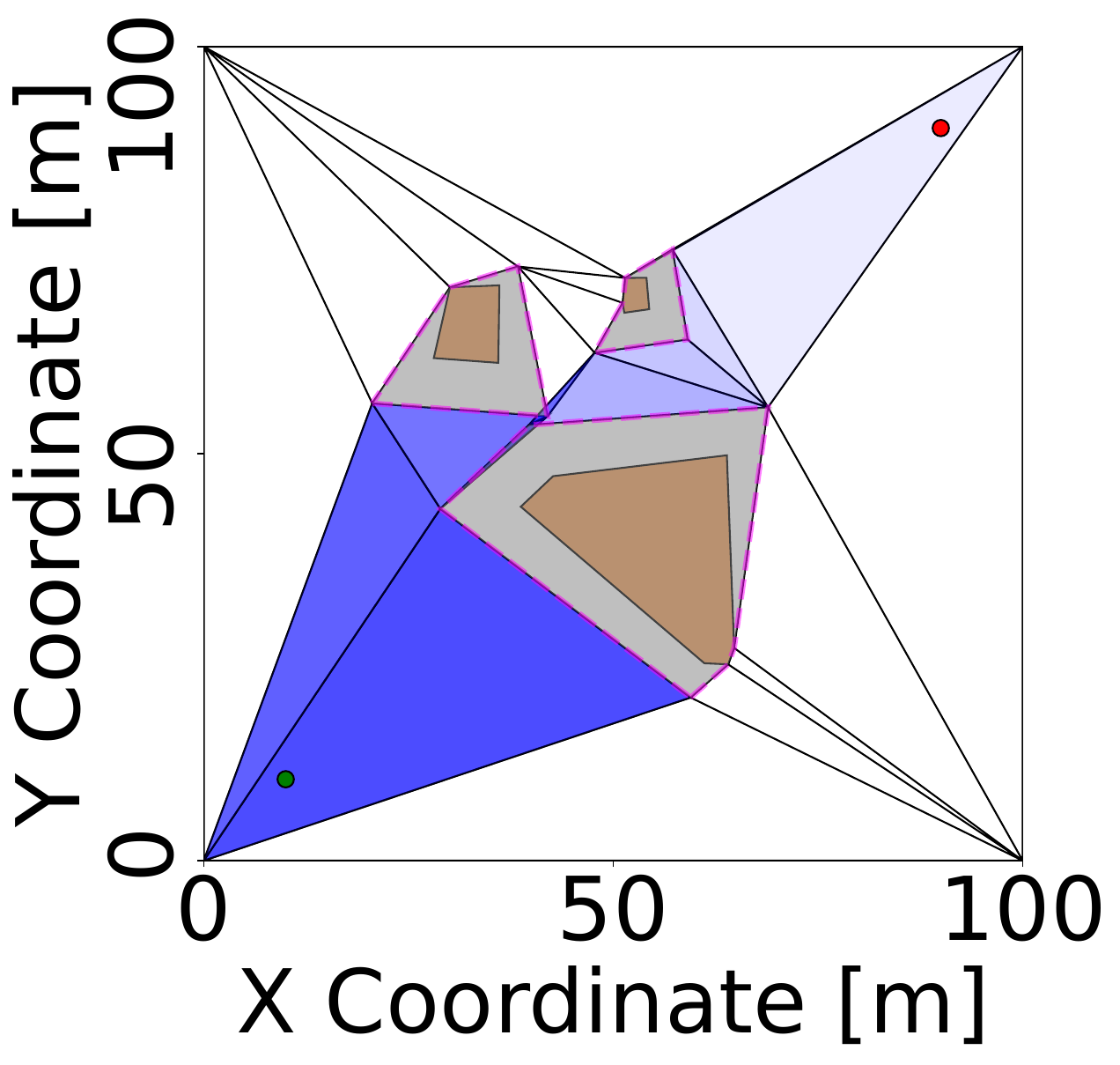}
         \end{subfigure}
    \end{minipage}
    ~
    \begin{minipage}[t]{0.45\columnwidth}
        \begin{subfigure}[b]{\textwidth}
           \includegraphics[width=\textwidth]{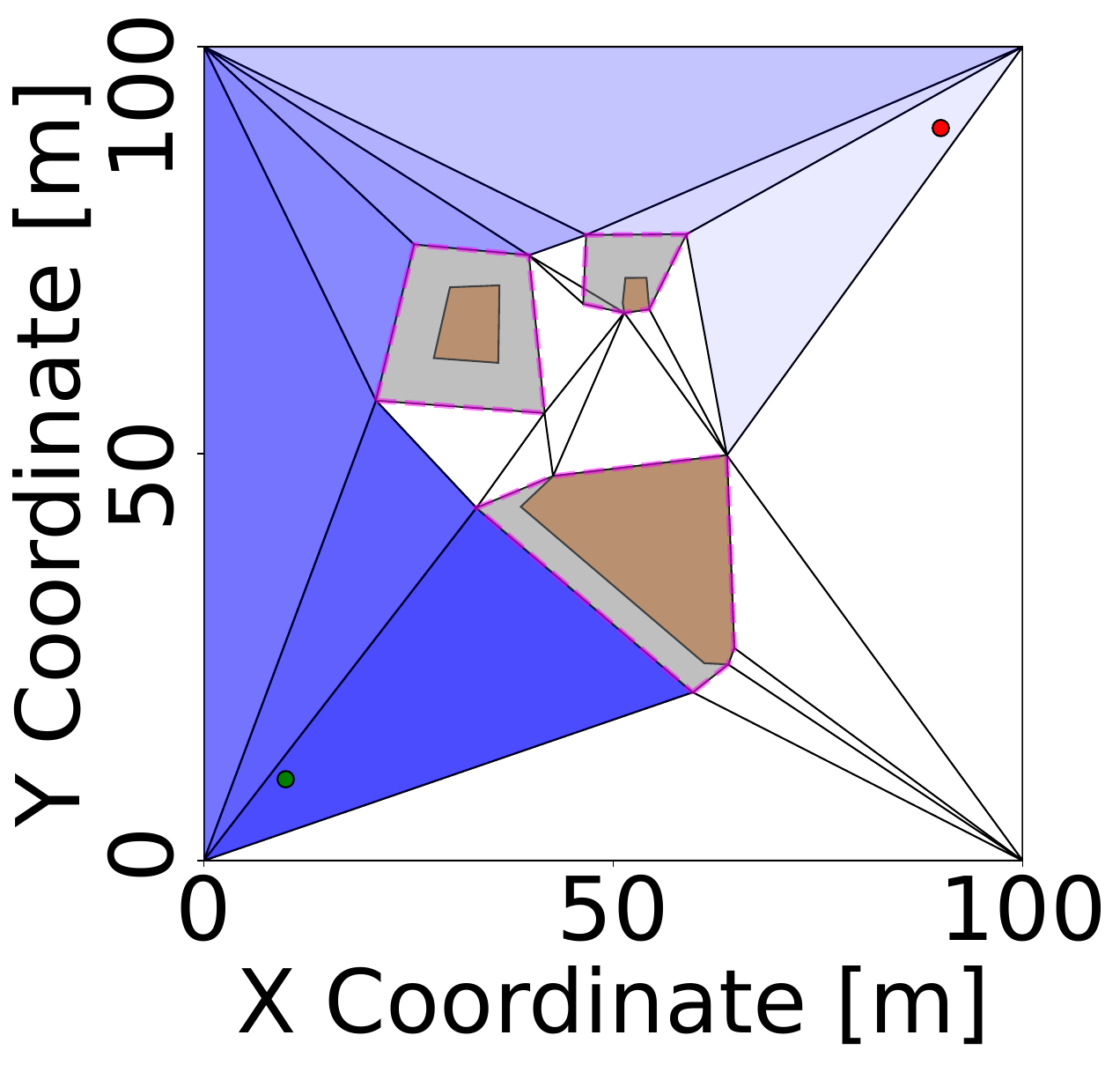}
         \end{subfigure}
    \end{minipage}
    \caption{Adaptive padding (gray) along a homotopy channel (blue triangles).  (\textbf{\textit{left}}) infeasible homotopy class due to the passage blocked by padding; and (\textbf{\textit{right}}) feasible.}
    \label{fig:adaptive-padding}
\end{figure}

Furthermore, external disturbances are not constant within a triangle. To capture local variability, we resample based on the standard deviation of the current’s direction and magnitude (Figure~\ref{fig:turning-circle} (mid)). Using these samples, we compute the percentile of distances to the closest edge and define a probabilistic collision bound $\sigma$, as shown in Figure~\ref{fig:cdf-best-effort}. This distance is used as a padding offset from the original CDT boundaries. In other words, the robot avoids collision with the no go zone with probability $\sigma$, implying a risk level of $1 - \sigma$. For a given homotopic channel, we apply padding along the channel. Our method adapts to both the direction of the external force and the obstacle’s relative location, allowing for a more precise estimation of safety margins (Figure~\ref{fig:turning-circle} (right)). Moreover, as shown in Figure~\ref{fig:adaptive-padding}, even the same constrained edge may experience different padding offsets depending on the homotopic options that change the relationship by the currents, vehicle, and constrained edge.

\subsection{High-level Planner}
Once we identify $k$ topologically distinct homotopic channels and apply adaptive no go zone padding, the high-level planner evaluates fuel efficiency  to select the optimal homotopy. %
To enable fair comparison across topologically distinct options, we first generate $N$ paths per homotopy class. For each homotopy, we sample points along the \textit{passing edges}---the shared boundaries between consecutive triangles within the homotopic sequence. The connection of these sampled points forms a complete path from the start to the goal. The convexity property of the CDT ensures that any straight-line connection between two points lying within or on a triangle is feasible---i.e., collision-free---thus guaranteeing that the full path from start to goal is geometrically valid.

However, geometric feasibility does not guarantee \textit{kinematic feasibility}. To ensure that each path can be smoothed into a kinematically feasible trajectory, we incorporate a check during sampling. Instead of relying on a fixed turning radius $r = v_{\text{thrust}} / \omega$, we introduce an effective radius $r' = v'_{\text{world}} / \omega$, where $v'_{\text{world}}$ represents the net effective speed based on average external disturbances within the local region $a$ around each waypoint. This accounts for variations in turning capability due to environmental forces.

We then evaluate each path within every homotopy by computing a fuel cost defined as:
\begin{equation} \label{eq:fuel_cost}
F = \int_{C} \alpha \cdot v_{thrust}^k \cdot \frac{ds}{v_{thrust} + \vec{c} \cdot \hat{T}(s)}
\end{equation}
where $\alpha$ is a system-dependent efficiency constant, 
$v_{thrust}$ is the constant thrust speed in still water, 
$k$ is a model exponent (typically \( k = 2 \) for quadratic drag), 
$ds$ is the differential path length, 
\( \vec{w} \) is the external current vector, and 
\( \hat{T}(s) \) is the unit tangent vector of the path at position \( s \). The dot product $\vec{c} \cdot \hat{T}(s)$ captures the component of the current aiding or opposing the motion.

Finally, the best homotopy class $h^\star$ minimizing the %
fuel cost is computed as the harmonic mean over sampled paths:
\begin{equation}
    h^\star = \arg\min_{h \in \mathcal{H}} \left( \left[ \frac{1}{|\mathcal{P}_h|} \sum_{i=1}^{|\mathcal{P}_h|} \frac{1}{F_{h,i}} \right]^{-1} \right)
\end{equation}
where \( \mathcal{H} \) is the set of all candidate homotopy classes considered by the planner; \( \mathcal{P}_h \) is the set of all valid paths in homotopy class \( h \); \( |\mathcal{P}_h| \) denotes the number of paths within \( \mathcal{P}_h \); and \( F_{h,i} \) is the total fuel consumption of the \( i \)-th path in homotopy class \( h \). The harmonic mean penalizes outliers with high fuel costs more than the arithmetic mean, promoting robustness in the selection process.

\subsection{Low-level Planner}

Once the optimal homotopy $h^\star$ is found, we proceed to determine the optimal path within the set $\mathcal{P}_{h^\star}$. Given the computations in the previous stage, each candidate path in $\mathcal{P}_{h^\star}$ satisfies kinematic feasibility under disturbances. We select the optimal path $p^\star$  that minimizes the fuel cost:
\begin{equation}
    p^\star = \arg\min_{p \in \mathcal{P}_{h^\star}} F(p)    
\end{equation}
Thus, the robot can follow the optimal trajectory while accounting for dynamic constraints and external disturbances, such as water currents, along the chosen homotopic channel.

\section{Experiments and Results}

We validated our method in real-world and simulated environments against \gridastar and sampling-based planners, including \rrt, \rrtstar, \prm. The suffix ``-D'' (e.g., \rrtdubins) denotes the integration of Dubins constraints.

For sampling-based methods, %
we performed 10 runs in each environment using different random seeds. For \prm, \prmdubins, and \gridastar, we extracted multiple topologically distinct paths using the $\mathcal{H}$-signature method proposed in~\cite{bhattacharya2012topological}, and used A$^\star$ as search algorithm. Therefore, \prm, \prmdubins, and \gridastar~naturally find the shortest path belonging to each homotopy class. Additionally, we applied the path smoothing technique \textsc{SimplifyMax} from the OMPL library~\cite{sucan2012the-open-motion-planning-library} to enhance kinematic feasibility—especially for methods that produce piecewise-linear paths such as \rrt, \rrtstar, and \prm. For \gridastar, we implemented Dubins-style motion with a fixed turning radius $r = v_{\text{thrust}} / \omega$~\cite{robot-turning-circle-2024}.

We used a vehicle model based on our custom-built ASV, which has a length of \SI{2}{\meter} and maximum linear and angular speeds of \SI{1.0}{\meter/\second} and \SI{35}{\degree/\second}, respectively. These parameters were obtained from real-world experiments, as described in~\cite{jeong-2020-iros}. We tested the vehicle motion under Dubins constraints during the experiments.

We compared performance across several metrics. \textit{Fuel}  is defined by Equation~\ref{eq:fuel_cost}.  \textit{Safety} is measured as the minimum distance to obstacles. The \textit{State} metric is the number of states explored by the planner. %
For fair computation comparison, we report only the \textit{State} metric. One example for insights on our method computational runtime: our method identifies homotopy classes within \SI{10}{\second} in an environment with 7 obstacles, and completes full path optimization using 500 samples per homotopy in approximately \SI{300}{\second}. 

While we report the \textit{Length} of each path for reference, our primary objective is not to minimize it. Instead, \textit{F/D} (fuel per unit distance) metric offers a more meaningful measure of path efficiency under external disturbances. 

Next, we discuss results; full data for all environments and additional experiments are in our GitHub repository.
\begin{table*}[t!]
\scriptsize
\centering
\renewcommand{\arraystretch}{0.9}
\renewcommand{\tabcolsep}{1.8pt}
\begin{tabular}{l|c|c|cc|cc|cc|cc|cc|cc|cc}
\toprule
& & \textbf{Ours} & \multicolumn{2}{c|}{\textbf{\rrt}} & \multicolumn{2}{c|}{\textbf{\rrtdubins}} & \multicolumn{2}{c|}{\textbf{\rrtstar}} & \multicolumn{2}{c|}{\textbf{\rrtstardubins}} & \multicolumn{2}{c|}{\textbf{\prm}} & \multicolumn{2}{c|}{\textbf{\prmdubins}} & \multicolumn{2}{c}{\textbf{\gridastar}} \\
\cmidrule(lr){3-3} \cmidrule(lr){4-5} \cmidrule(lr){6-7} \cmidrule(lr){8-9} \cmidrule(lr){10-11} \cmidrule(lr){12-13} \cmidrule(lr){14-15} \cmidrule(lr){16-17}
\textbf{Env.} & \textbf{Metric} & \textbf{S} & \textbf{O} & \textbf{S} & \textbf{O} & \textbf{S} & \textbf{O} & \textbf{S} & \textbf{O} & \textbf{S} & \textbf{O} & \textbf{S} & \textbf{O} & \textbf{S} & \textbf{O} & \textbf{S} \\
\midrule
\multirow{10}{*}{\parbox{1.35cm}{\textbf{4-Gyre} \\ (200$\times$200) \\ k=16}} & \multirow{2}{*}{Fuel $\downarrow$} & \cellcolor{green!30}\textbf{202.25} & 310.57 & 236.27 & 377.45 & 293.34 & 288.07 & 235.09 & 354.94 & 269.92 & 295.35 & 219.17 & 657.11 & 224.84 & 225.68 & \cellcolor{yellow!30}\textbf{215.15} \\
& & \cellcolor{green!30}-- & $\pm$77.59 & $\pm$39.47 & $\pm$93.25 & $\pm$83.49 & $\pm$65.32 & $\pm$38.15 & $\pm$81.31 & $\pm$65.90 & $\pm$25.13 & $\pm$20.30 & $\pm$64.85 & $\pm$75.90 & -- & \cellcolor{yellow!30}-- \\
\cline{2-17}
& \multirow{2}{*}{Safety $\uparrow$} & \cellcolor{green!30}\textbf{16.40} & 1.77 & 1.18 & 3.38 & 4.00 & 1.77 & 1.05 & 3.23 & 3.01 & 2.64 & 0.99 & 3.98 & \cellcolor{yellow!30}\textbf{4.20} & 1.42 & 0.97 \\
& & \cellcolor{green!30}-- & $\pm$0.84 & $\pm$0.48 & $\pm$2.93 & $\pm$2.82 & $\pm$0.85 & $\pm$0.47 & $\pm$2.93 & $\pm$2.66 & $\pm$1.57 & $\pm$0.44 & $\pm$5.37 & \cellcolor{yellow!30}$\pm$4.14 & -- & -- \\
\cline{2-17}
& \multirow{2}{*}{F/D $\downarrow$} & \cellcolor{green!30}\textbf{0.680} & 0.99 & 0.92 & 1.19 & 1.13 & 1.01 & 0.93 & 1.14 & 1.15 & 0.83 & 0.86 & 1.04 & 0.86 & 0.857 & \cellcolor{yellow!30}\textbf{0.841} \\
& & \cellcolor{green!30}-- & $\pm$0.22 & $\pm$0.15 & $\pm$0.28 & $\pm$0.32 & $\pm$0.18 & $\pm$0.15 & $\pm$0.25 & $\pm$0.28 & $\pm$0.07 & $\pm$0.06 & $\pm$0.07 & $\pm$0.27 & -- & \cellcolor{yellow!30}-- \\
\cline{2-17}
& States $\downarrow$ & \cellcolor{green!30}\textbf{86} & \multicolumn{2}{c|}{\cellcolor{yellow!30}\textbf{394.9$\pm$75.2}} & \multicolumn{2}{c|}{476.7$\pm$127.8} & \multicolumn{2}{c|}{\cellcolor{yellow!30}\textbf{394.9$\pm$75.2}} & \multicolumn{2}{c|}{526.5$\pm$135.5} & \multicolumn{2}{c|}{17662$\pm$5275} & \multicolumn{2}{c|}{19174$\pm$7069} & \multicolumn{2}{c}{86521} \\
& & \cellcolor{green!30}(26) & \multicolumn{2}{c|}{} & \multicolumn{2}{c|}{} & \multicolumn{2}{c|}{} & \multicolumn{2}{c|}{} & \multicolumn{2}{c|}{(2000)} & \multicolumn{2}{c|}{(2000)} & \multicolumn{2}{c}{(10000)} \\
\cline{2-17}
& \multirow{2}{*}{Length*} & 297.59 & 311.28 & 256.12 & 317.14 & 259.72 & 305.71 & 255.84 & 309.09 & 257.59 & 291.88 & 255.99 & 358.17 & 266.98 & 263.25 & 255.92 \\
& & (--) & $\pm$10.54 & $\pm$0.42 & $\pm$21.21 & $\pm$3.65 & $\pm$8.72 & $\pm$0.43 & $\pm$17.81 & $\pm$2.52 & $\pm$7.95 & $\pm$0.89 & $\pm$28.03 & $\pm$4.17 & -- & -- \\
\midrule
\multirow{10}{*}{\parbox{1.35cm}{\textbf{Hansando} \\ (1533$\times$1619) \\ k=10}} & \multirow{2}{*}{Fuel $\downarrow$} & \cellcolor{green!30}\textbf{2639.39} & 3566.74 & 3957.74 & 3634.23 & 3618.30 & 3564.63 & 3958.51 & 3428.69 & 3270.21 & 3148.71 & 2854.45 & 3470.89 & 2844.51 & 2863.82 & \cellcolor{yellow!30}\textbf{2838.78} \\
& & \cellcolor{green!30}-- & $\pm$159.10 & $\pm$618.34 & $\pm$355.33 & $\pm$421.82 & $\pm$159.10 & $\pm$618.35 & $\pm$170.68 & $\pm$388.83 & $\pm$183.73 & $\pm$204.84 & $\pm$96.70 & $\pm$36.09 & -- & \cellcolor{yellow!30}-- \\
\cline{2-17}
& \multirow{2}{*}{Safety $\uparrow$} & \cellcolor{yellow!30}\textbf{22.61} & 8.21 & 1.28 & 4.84 & 15.36 & 8.21 & 1.28 & 8.53 & \cellcolor{green!30}\textbf{28.65} & 4.15 & 2.33 & 3.44 & 2.80 & 1.18 & 1.46 \\
& & \cellcolor{yellow!30}-- & $\pm$13.41 & $\pm$0.32 & $\pm$5.76 & $\pm$12.79 & $\pm$13.41 & $\pm$0.32 & $\pm$10.38 & \cellcolor{green!30}$\pm$13.76 & $\pm$4.90 & $\pm$2.01 & $\pm$2.27 & $\pm$2.68 & -- & -- \\
\cline{2-17}
& \multirow{2}{*}{F/D $\downarrow$} & \cellcolor{green!30}\textbf{0.951} & 1.41 & 2.02 & 1.45 & 1.82 & 1.41 & 2.02 & 1.44 & 1.90 & \cellcolor{yellow!30}\textbf{1.03} & 1.14 & 1.26 & 1.34 & 1.092 & 1.124 \\
& & \cellcolor{green!30}-- & $\pm$0.14 & $\pm$0.43 & $\pm$0.16 & $\pm$0.25 & $\pm$0.14 & $\pm$0.43 & $\pm$0.15 & $\pm$0.25 & \cellcolor{yellow!30}$\pm$0.06 & $\pm$0.08 & $\pm$0.13 & $\pm$0.13 & -- & -- \\
\cline{2-17}
& States $\downarrow$ & \cellcolor{green!30}\textbf{161} & \multicolumn{2}{c|}{2076.7$\pm$810.2} & \multicolumn{2}{c|}{2120.9$\pm$601.8} & \multicolumn{2}{c|}{2076.7$\pm$810.2} & \multicolumn{2}{c|}{\cellcolor{yellow!30}\textbf{1831.6$\pm$539.3}} & \multicolumn{2}{c|}{21514.1$\pm$6425} & \multicolumn{2}{c|}{21692.1$\pm$1937} & \multicolumn{2}{c}{42847} \\
& & \cellcolor{green!30}(34) & \multicolumn{2}{c|}{} & \multicolumn{2}{c|}{} & \multicolumn{2}{c|}{} & \multicolumn{2}{c|}{} & \multicolumn{2}{c|}{(7000)} & \multicolumn{2}{c|}{(7000)} & \multicolumn{2}{c}{(17408)} \\
\cline{2-17}
& \multirow{2}{*}{Length*} & 2775.04 & 2555.30 & 1997.28 & 2509.89 & 1992.01 & 2555.30 & 1997.28 & 2404.90 & 1935.47 & 3041.34 & 2560.38 & 2703.50 & 2095.76 & 2623.33 & 2526.09 \\
& & -- & $\pm$251.35 & $\pm$204.54 & $\pm$102.09 & $\pm$63.87 & $\pm$251.35 & $\pm$204.54 & $\pm$120.28 & $\pm$146.81 & $\pm$228.30 & $\pm$160.12 & $\pm$154.31 & $\pm$134.86 & -- & -- \\
\midrule
\multirow{10}{*}{\parbox{1.35cm}{\textbf{Far East} \\ (1360$\times$1270) \\ k=10}} & \multirow{2}{*}{Fuel $\downarrow$} & \cellcolor{green!30}\textbf{1247.83} & 1957.68 & 1512.88 & 1899.56 & 1472.81 & 1956.54 & 1513.44 & 1859.15 & 1447.35 & 1569.65 & 1386.34 & 1760.51 & 1387.14 & 1479.12 & \cellcolor{yellow!30}\textbf{1288.27} \\
& & \cellcolor{green!30}-- & $\pm$188.16 & $\pm$14.07 & $\pm$123.32 & $\pm$111.28 & $\pm$188.16 & $\pm$14.07 & $\pm$143.95 & $\pm$125.91 & $\pm$129.47 & $\pm$117.33 & $\pm$46.64 & $\pm$73.32 & -- & \cellcolor{yellow!30}-- \\
\cline{2-17}
& \multirow{2}{*}{Safety $\uparrow$} & \cellcolor{green!30}\textbf{37.69} & 2.11 & 1.13 & 1.95 & 3.63 & 2.11 & 1.24 & 2.51 & \cellcolor{yellow!30}\textbf{4.56} & 4.39 & 1.25 & 3.12 & 2.93 & 1.05 & 1.80 \\
& & \cellcolor{green!30}-- & $\pm$1.10 & $\pm$0.12 & $\pm$1.27 & $\pm$2.56 & $\pm$1.10 & $\pm$0.12 & $\pm$2.00 & \cellcolor{yellow!30}$\pm$3.05 & $\pm$3.29 & $\pm$0.92 & $\pm$3.02 & $\pm$3.44 & -- & -- \\
\cline{2-17}
& \multirow{2}{*}{F/D $\downarrow$} & \cellcolor{green!30}\textbf{0.776} & 0.95 & 0.95 & 0.93 & 0.90 & 0.95 & 0.95 & 0.93 & 0.92 & 0.94 & 0.88 & 0.88 & 0.85 & 0.889 & \cellcolor{yellow!30}\textbf{0.834} \\
& & \cellcolor{green!30}-- & $\pm$0.05 & $\pm$0.00 & $\pm$0.05 & $\pm$0.06 & $\pm$0.05 & $\pm$0.00 & $\pm$0.07 & $\pm$0.06 & $\pm$0.08 & $\pm$0.07 & $\pm$0.03 & $\pm$0.06 & -- & \cellcolor{yellow!30}-- \\
\cline{2-17}
& States $\downarrow$ & \cellcolor{green!30}\textbf{183} & \multicolumn{2}{c|}{2628.9$\pm$1318.1} & \multicolumn{2}{c|}{\cellcolor{yellow!30}\textbf{2360.5$\pm$663.5}} & \multicolumn{2}{c|}{2628.9$\pm$1318.1} & \multicolumn{2}{c|}{2451.6$\pm$710.9} & \multicolumn{2}{c|}{57335.3$\pm$12757} & \multicolumn{2}{c|}{56395.4$\pm$11991} & \multicolumn{2}{c}{32794} \\
& & \cellcolor{green!30}(73) & \multicolumn{2}{c|}{} & \multicolumn{2}{c|}{} & \multicolumn{2}{c|}{} & \multicolumn{2}{c|}{} & \multicolumn{2}{c|}{(7000)} & \multicolumn{2}{c|}{(7000)} & \multicolumn{2}{c}{(24948)} \\
\cline{2-17}
& \multirow{2}{*}{Length*} & 1607.33 & 2051.72 & 1587.42 & 2037.12 & 1638.50 & 2051.72 & 1587.42 & 1971.90 & 1613.50 & 1706.03 & 1544.24 & 1879.85 & 1547.43 & 1664.21 & 1545.30 \\
& & -- & $\pm$92.45 & $\pm$17.03 & $\pm$80.75 & $\pm$44.35 & $\pm$92.45 & $\pm$17.03 & $\pm$101.19 & $\pm$51.67 & $\pm$89.76 & $\pm$19.06 & $\pm$83.26 & $\pm$27.12 & -- & -- \\
\bottomrule
\end{tabular}%
\caption{
\setlength{\fboxsep}{0pt}
Performance comparison of paths across environments (size in meters). The best and second-best performances are highlighted in {\small \colorbox{green!30}{\textit{green}}} and {\small \colorbox{yellow!50}{\textit{yellow}}}. ``O'' and ``S'' denote original and smoothed paths. For sampling-based methods, values are mean in first row and ±std in second row over $10$ runs. The number in parentheses in the \textit{States} metric indicates the %
size of the discretized environment (Ours: triangulation vertices; \prm~and~\prmdubins: sampled nodes; \gridastar: grid cells with \SI{2}{\meter} resolution for 4-gyre, and \SI{10}{\meter} for Hansando and Far East). $k$ denotes the maximum number of homotopy classes considered. *Note: our objective is not to minimize path length, but we report it for reference.
}
\label{tab:performance_comparison}
\end{table*}

\subsection{Comparative Studies}
We conducted a comparative analysis for two scenarios involving different start--goal combinations, using the environment from~\cite{Kularatne-RSS-16}. 

\begin{figure}[h]
    \centering
    \begin{minipage}[t]{0.58\columnwidth}
        \begin{subfigure}[b]{\textwidth}
           \includegraphics[width=\textwidth]{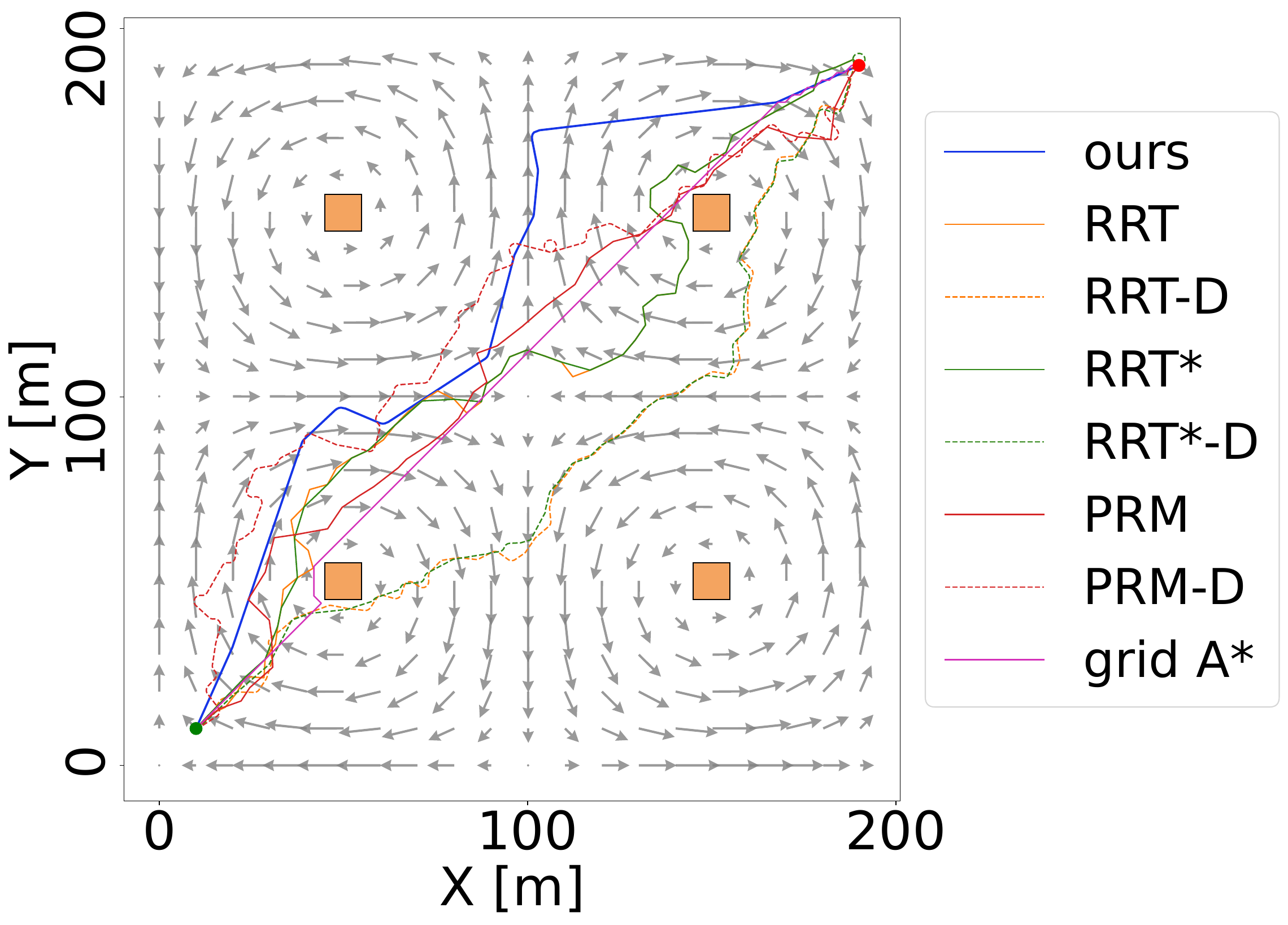}
         \end{subfigure}
    \end{minipage}
    ~
    \begin{minipage}[t]{0.37\columnwidth}
        \begin{subfigure}[b]{\textwidth}
           \includegraphics[width=\textwidth]{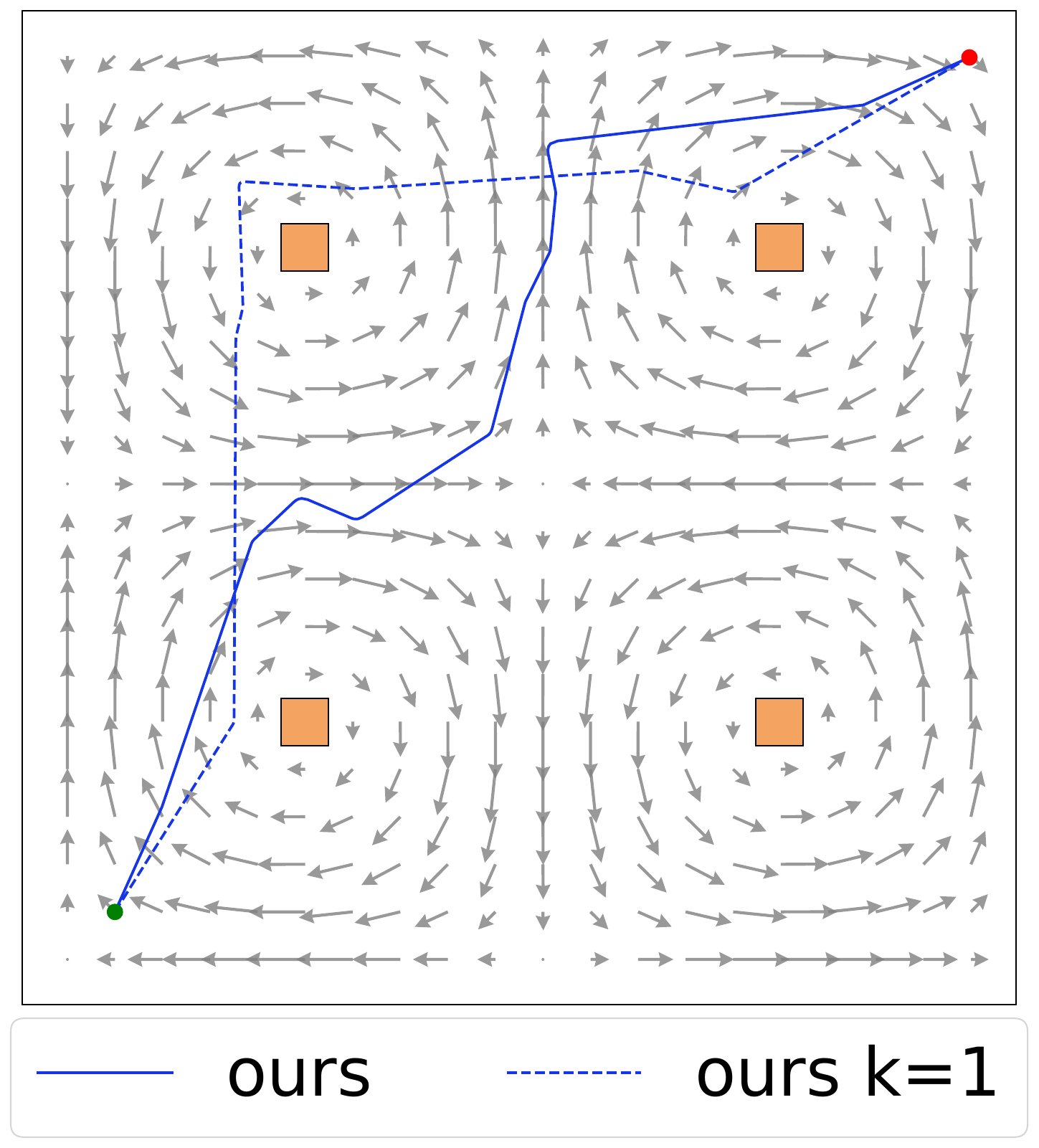}
         \end{subfigure}
    \end{minipage}
    \caption{Qualitative comparison of paths in 4 gyres (start: green, goal: red). (\textbf{\textit{left}}) ours and original paths by baseline methods; and (\textbf{\textit{right}}) ours method with $k=16$ vs. $k=1$.}
    \label{fig:4-core-comparison}
\end{figure}

Our proposed method successfully identifies an optimal path by leveraging favorable current directions (Figure~\ref{fig:4-core-comparison} (\textit{left})). Specifically, it identifies multiple topologically distinct options, selects the optimal homotopy class, and then chooses the best path within that class. As a result, the vehicle achieves reduced fuel consumption--even when the selected path is longer--by following routes where the currents provide assistance. See Table~\ref{tab:performance_comparison} for quantitative results.

We further evaluated our method under different values of $k$, the maximum number of homotopy classes considered—see Figure~\ref{fig:4-core-comparison} (\textit{right}). Notably, the case with $k = 16$ enabled the method to find the optimal path, with improvements compared to $k=1$: fuel consumption (202.25 vs. 325.18), safety (16.40 vs. 7.70), path length (297.59 vs. 322.80), and fuel efficiency (\textit{F/D}: 0.680 vs. 1.007).

\subsection{Real-world Environment}
We validated our proposed method in real-world scenarios, using nautical charts and sea surface current data from Copernicus~\cite{copernicus} and the National Ocean Satellite Center~\cite{national-ocean-sat}. The original nautical chart polylines were simplified using the Douglas--Peucker algorithm \cite{douglas1973algorithms}. Each environment was scaled down to ASV-relevant dimensions for simulation purposes while preserving the geometric structure.

We selected three 
distinct environments: %
(1) \textbf{Hansando} (1:15 scale), located in the southern waters of Korea, features a complex current field and intricate topological structure within a coastal area~\cite{nga2022sailing}, with generally strong adverse current against the navigation direction; 
(2) \textbf{Far East Asian waters} (1:2000 scale), which are consistently affected by the strong northeastward-flowing Kuroshio current, influencing nearby maritime traffic~\cite{kuroshio-ship-routing-2013}; and  
(3) the \textbf{Palawan Passage} (1:5000 scale), where current directions change significantly between seasons, introducing dynamic navigational challenges~\cite{southchinasea-2000}.

\begin{figure}[t!]
    \centering
    \begin{minipage}[t]{0.44\columnwidth}
        \begin{subfigure}[b]{\textwidth}
           \includegraphics[width=\textwidth]{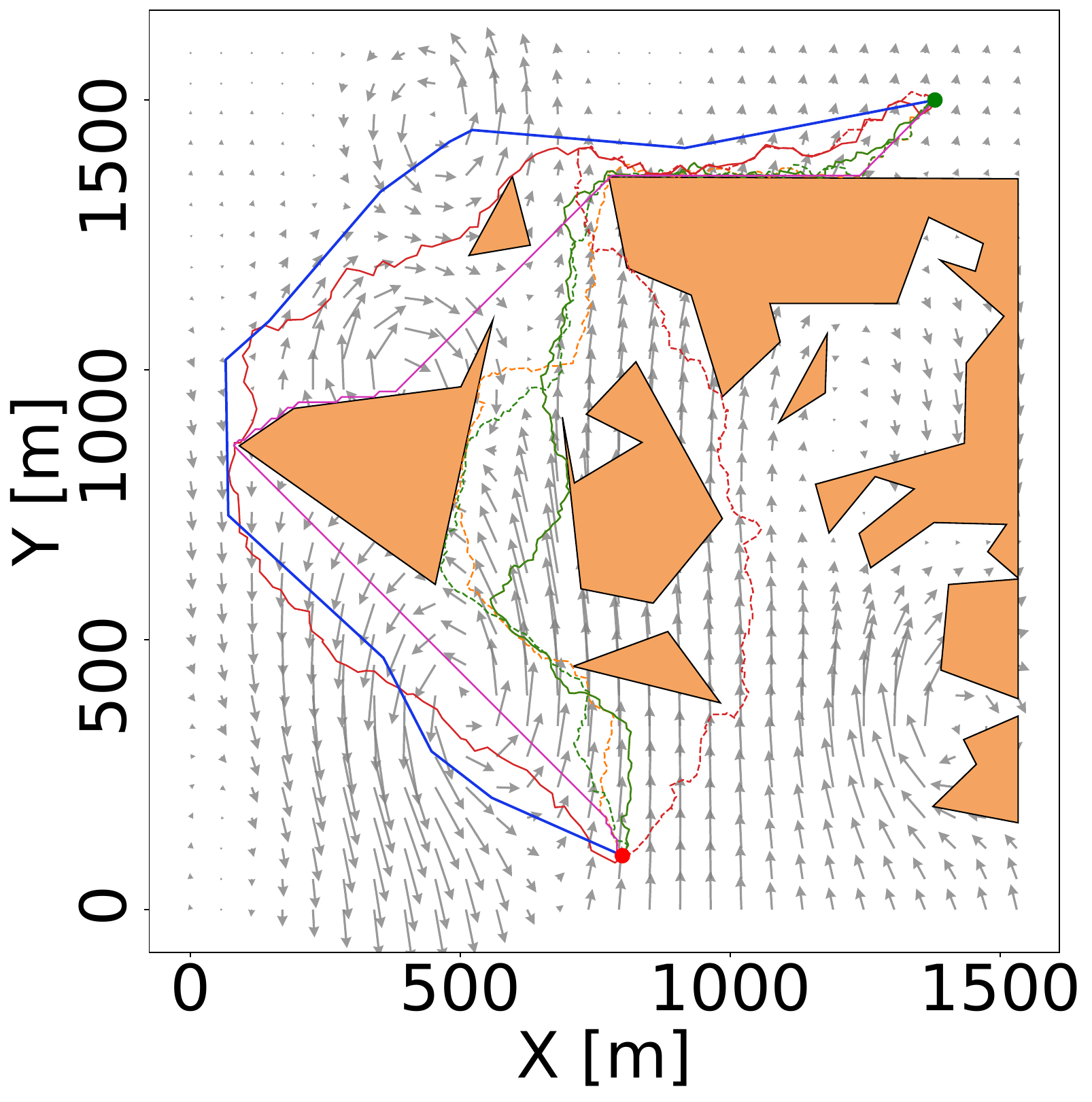}
         \end{subfigure}
    \end{minipage}
    ~

    \caption{Qualitative comparison using real current data in Hansando area with the sea surface current in summer 2024.} 
    \label{fig:real-combined}
\end{figure}

First, 
 although the path found by our method is not the shortest, the fuel efficiency and \textit{F/D} are the best (Figure~\ref{fig:real-combined} and Table~\ref{tab:performance_comparison}~(Hansando)). In some cases, the smoothed path is significantly worse than the original path, e.g., \rrt. This occurs as the smoothing process does not explicitly account for external disturbances during optimization, leading to paths that are adversely affected by opposing currents.

\begin{figure}[t]
    \centering
    \begin{subfigure}[b]{.40\columnwidth}
        \centering
        \includegraphics[width=\textwidth]{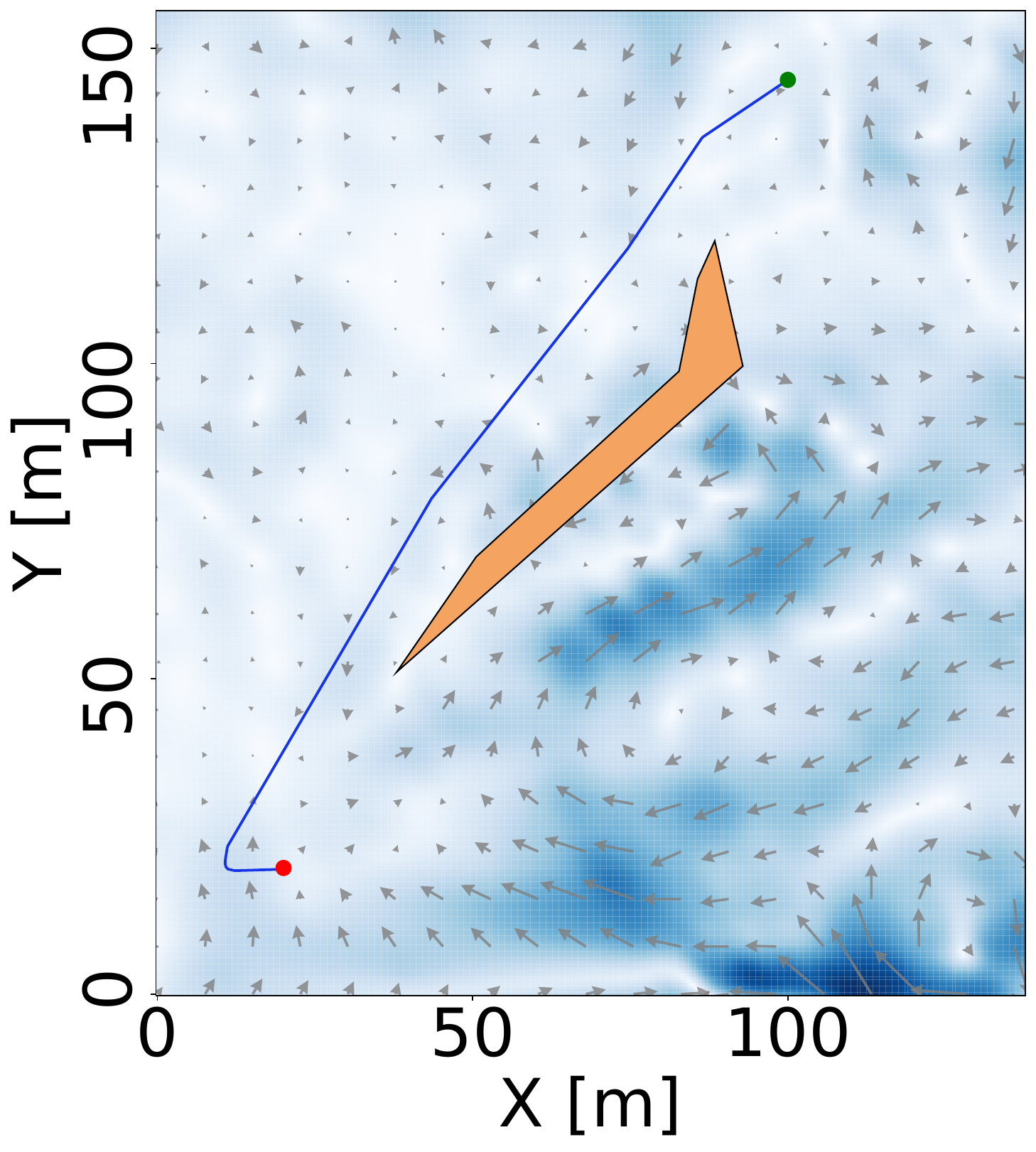}
    \end{subfigure}
    \centering
    \begin{subfigure}[b]{.52\columnwidth}
        \centering
        \includegraphics[width=\textwidth]{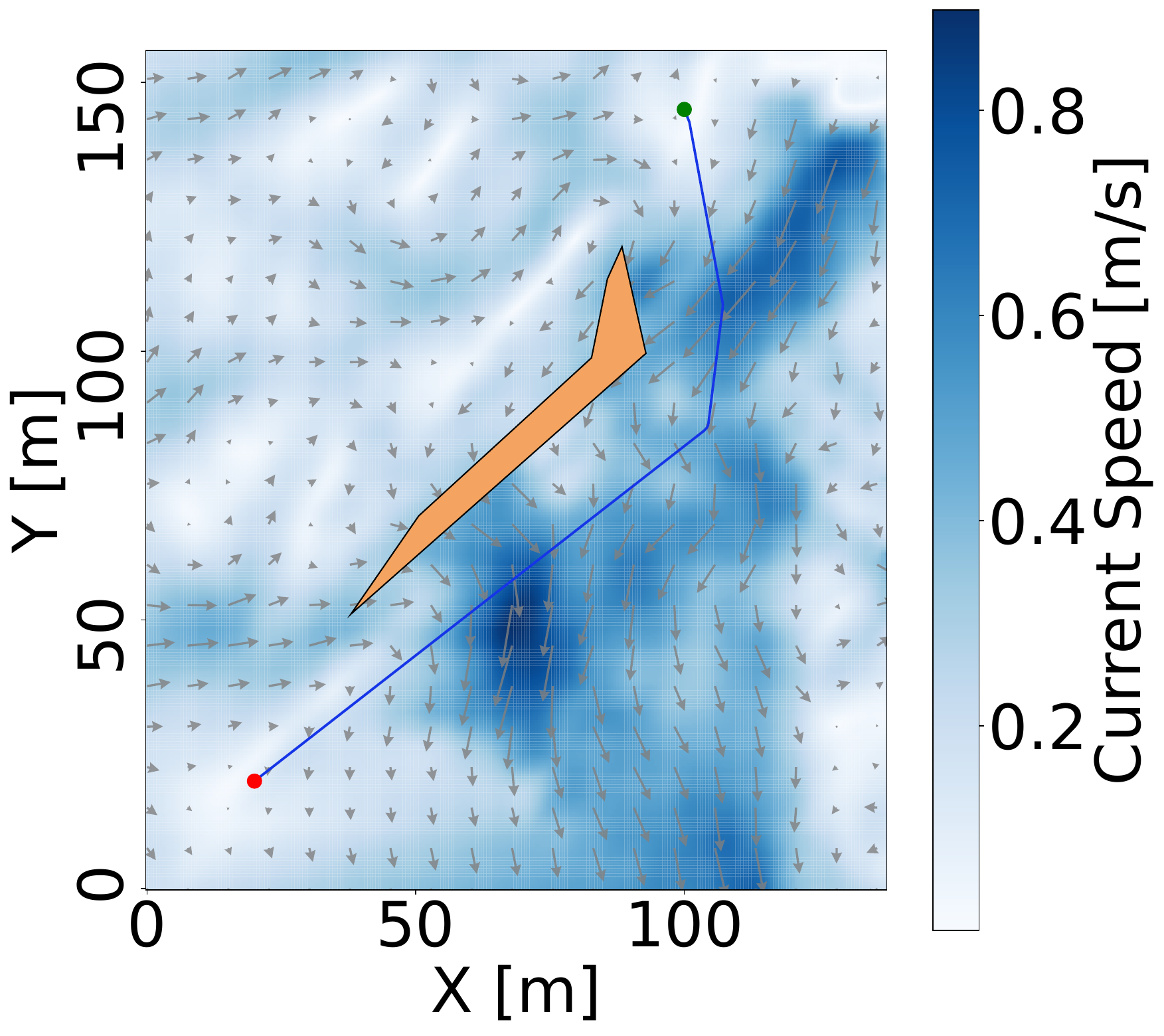}
    \end{subfigure}
    \caption{Qualitative comparison of paths proposed by our method in the waters near Palawan Passage in the Philippines. (\textbf{\textit{left}}) summer sea surface current in July 2024; and (\textbf{\textit{right}}) winter sea surface current in December 2024.}
    \label{fig:palawan-comparison}
\end{figure}

Next, we show that our method can adaptively choose the optimal path under varying environmental conditions, even with the same start and goal points—see Figure~\ref{fig:palawan-comparison} and Table~\ref{tab:palawan-comparison}. During the summer, our algorithm found the west-side route to avoid strong northeast-directed currents along the east side of Palawan Island. During winter instead, our method found an efficient route along the east side thanks to the southwest-directed currents. 

\begin{table}[]
    \scriptsize
    \centering
    \begin{tabular}{l|c|c|c|c}
    \hline
    \textbf{Ours} & \textbf{Fuel} & \textbf{Safety} & \textbf{Length} & \textbf{F/D} \\
    \hline
    Summer & 160.563 & 9.153 & 164.586 & 0.976 \\
    \hline
    Winter & 131.787 & 13.133 & 167.053 & 0.789 \\
    \hline
    \end{tabular}
    \caption{Performance comparison of paths across the summer and winter season.}
\label{tab:palawan-comparison}
\end{table}
\subsection{Ablation Studies}
\begin{figure}[t]
    \centering
    \begin{subfigure}[b]{.2\columnwidth}
        \centering
        \includegraphics[width=\textwidth]{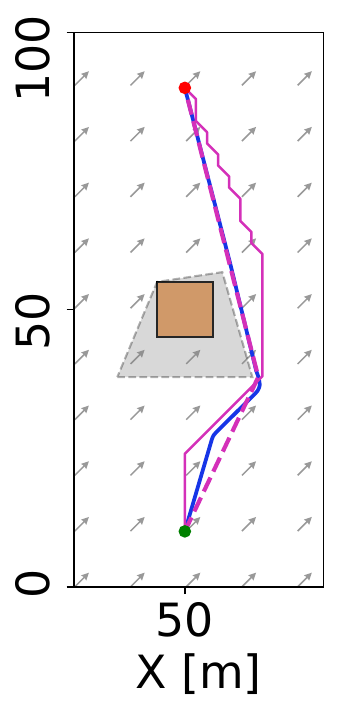}
    \end{subfigure}
    \begin{subfigure}[b]{.2\columnwidth}
        \centering
        \includegraphics[width=\textwidth]{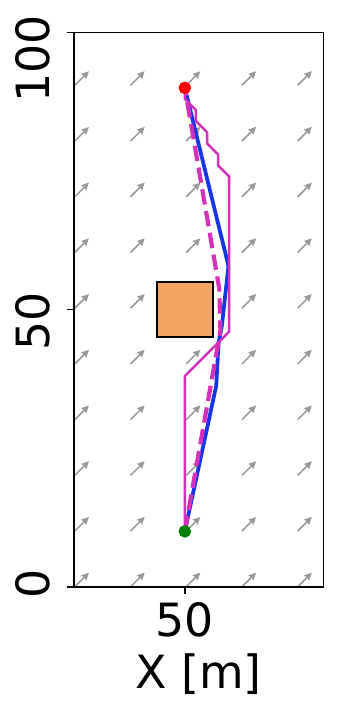}
    \end{subfigure}
    \begin{subfigure}[b]{.2\columnwidth}
        \centering
        \includegraphics[width=\textwidth]{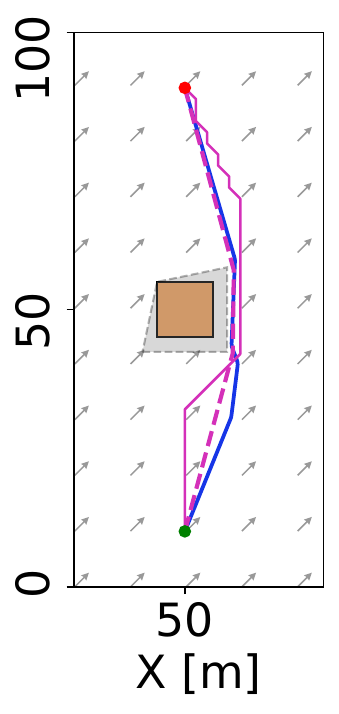}
    \end{subfigure}
    \begin{subfigure}[b]{.2\columnwidth}
        \centering
        \includegraphics[width=\textwidth]{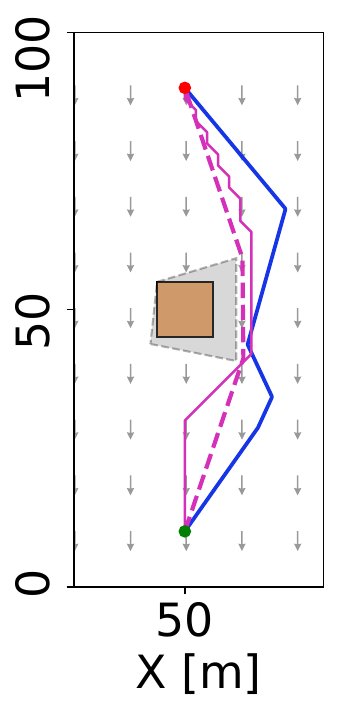}
    \end{subfigure}
    \caption{Qualitative comparison of paths by ablation tests with the current direction $\beta$. The padded area is colored in \textit{gray} and the original obstacle area is in \textit{brown}. (\textbf{\textit{left}}) adaptive padding ($\beta$=\SI{45}{\degree}); (\textbf{\textit{center-left}}) no padding ($\beta$=\SI{45}{\degree}); (\textbf{\textit{center-right}}) fixed padding ($\beta$=\SI{45}{\degree}); (\textbf{\textit{right}}) adaptive padding ($\beta$=\SI{-90}{\degree}).}
    \label{fig:ablation}
\end{figure}

We investigated how padding schemes, including our adaptive padding, and the path planning algorithms behave in controlled scenarios. We confined the homotopy class (i.e., the path must pass on the right side of the obstacle) and varied the direction of a uniform water current field (Figure~\ref{fig:ablation}).

When the smoothed shortest path aligns with the optimal one, \gridastar\ and our method yield comparable fuel efficiency, provided they share the same padding scheme (\textit{adaptive}, \textit{no}, or \textit{fixed padding}).
Notably, while \SI{2}{\meter} resolution \gridastar\ initially appears safer, smoothing results in tighter paths around obstacles, ultimately bringing its safety performance in line with our approach.

In contrast, there are cases where our method significantly outperforms \gridastar. %
Our planner produced longer paths that are more fuel-efficient, especially when the smoothed shortest path encounters opposing currents that increase fuel consumption. Our method generates zig-zag paths with longer distances, but successfully avoids adverse currents and reduces overall fuel cost (Figures~\ref{fig:ablation}(right) and Table~\ref{tab:ablation_results}~(case d)). In these scenarios, \gridastar\ selects shorter but less efficient paths under the same homotopy and padding scheme.

Lastly, Figures~\ref{fig:ablation}(left) and~\ref{fig:ablation}(right) demonstrate how our method ensures safety by satisfying a probabilistic collision bound through adaptive padding during passage, in contrast to the fixed padding shown in Figure~\ref{fig:ablation}(center-right).

\begin{table}[]
\scriptsize
\centering
\renewcommand{\arraystretch}{0.9}
\begin{tabular}{@{}lcc|ccc@{}}
\toprule
\textbf{Case} & \textbf{Type} & \textbf{Metric} & \textbf{Ours} & \textbf{\gridastar-O} & \textbf{\gridastar-S} \\
\midrule
\multirow{4}{*}{a} 
& \multirow{4}{*}{\begin{tabular}{@{}c@{}}Adaptive\\(45°)\end{tabular}}
& Fuel $\downarrow$ & 56.76 & 63.77 & 55.72 \\
& & Safety $\uparrow$ & 3.87 & 9.00 & 3.77 \\
& & F/D $\downarrow$ & 0.66 & 0.70 & 0.66 \\
& & Length* & 85.80 & 91.60 & 84.65 \\
\midrule
\multirow{4}{*}{b} 
& \multirow{4}{*}{\begin{tabular}{@{}c@{}}No\\(45°)\end{tabular}}
& Fuel $\downarrow$ & 52.86 & 58.34 & 52.23 \\
& & Safety $\uparrow$ & 1.26 & 1.42 & 1.00 \\
& & F/D $\downarrow$ & 0.65 & 0.67 & 0.64 \\
& & Length* & 81.68 & 86.63 & 81.10 \\
\midrule
\multirow{4}{*}{c} 
& \multirow{4}{*}{\begin{tabular}{@{}c@{}}Fixed\\(45°)\end{tabular}}
& Fuel $\downarrow$ & 54.39 & 60.14 & 53.50 \\
& & Safety $\uparrow$ & 3.48 & 5.00 & 3.69 \\
& & F/D $\downarrow$ & 0.65 & 0.68 & 0.65 \\
& & Length* & 83.19 & 88.28 & 82.31 \\
\midrule
\multirow{4}{*}{d} 
& \multirow{4}{*}{\begin{tabular}{@{}c@{}}Adaptive\\(270°)\end{tabular}}
& Fuel $\downarrow$ & 311.29 & 357.97 & 360.37 \\
& & Safety $\uparrow$ & 6.46 & 7.00 & 5.43 \\
& & F/D $\downarrow$ & 3.34 & 3.98 & 4.32 \\
& & Length* & 93.16 & 89.94 & 83.45 \\
\bottomrule
\end{tabular}%
\caption{Ablation study results. *Note: our objective is not to minimize path length as noted in Table~\ref{tab:performance_comparison}.}
\label{tab:ablation_results}
\end{table}

\subsection{Contingency Maneuver}
We validated how our proposed method ensures the avoidance of ICS, i.e., guaranteeing no collision even after executing a contingency maneuver: based on the paths found, %
we conducted forward simulations assuming the presence of an unexpected obstacle %
every \SI{1}{\meter} along the path, and performed a contingency maneuver to test safety.

Under dynamic disturbances, our proposed method consistently avoided collisions (Table~\ref{tab:contingency_result}). In contrast, state-of-the-art approaches experienced collisions---an outcome that is critical in real-world scenarios where the uncertainty of environmental disturbances poses significant safety risks.

\begin{table}[t]
\scriptsize
\renewcommand{\tabcolsep}{1.5pt}
\renewcommand{\arraystretch}{0.9}
\centering
\begin{tabular}{llrrrrrrrr}
\toprule
\textbf{Scenario} & \textbf{Metric} & \textbf{Ours} & \textbf{\rrt} & \textbf{\rrtdubins} & \textbf{\rrtstar} & \textbf{\rrtstardubins} & \textbf{\prm} & \textbf{\prmdubins} & \textbf{\gridastar} \\
\midrule
\multirow{2}{*}{4-gyre}    & Collision & \cellcolor{green!20}0 & 79 & 21 & \cellcolor{red!20}81 & 30 & 69 & 23 & 79 \\
                           & Trials    & 1494 & 1280 & 1288 & 1280 & 1285 & 1280 & 1336 & 1280 \\
\midrule
\multirow{2}{*}{Hansando}  & Collision & \cellcolor{green!20}0 & 34 & 8 & \cellcolor{red!20}61 & 15 & 14 & 19 & 61 \\
                           & Trials    & 2788 & 1906 & 1995 & 1907 & 2025 & 2561 & 2096 & 2527 \\
\midrule
\multirow{2}{*}{Far East}  & Collision & \cellcolor{green!20}0 & \cellcolor{red!20}72 & \cellcolor{green!20}0 & 71 & \cellcolor{green!20}0 & 71 & 20 & 31 \\
                           & Trials    & 1608 & 1545 & 1573 & 1545 & 1582 & 1545 & 1548 & 1546 \\
\midrule
\multirow{2}{*}{Total}     & Collision & \cellcolor{green!20}0 & 185 & 29 & \cellcolor{red!20}213 & 45 & 154 & 62 & 128 \\
                           & Trials    & 5890 & 4731 & 4856 & 4732 & 4892 & 5386 & 4980 & 5353 \\
\bottomrule
\end{tabular}%
\caption{
\setlength{\fboxsep}{0pt}
Collision counts and total attempts during the contingency maneuvers over the found paths. {\small \colorbox{green!20}{\textit{green}}}: the lowest collision count; {\small \colorbox{red!20}{\textit{red}}}: the highest count.}
\label{tab:contingency_result}
\end{table}

\section{Conclusion and Future Steps}\label{sec:conclusion}
We presented a risk- and energy-aware global planner for ASVs navigating dynamic disturbances. By integrating adaptive padding and a worst-case best-effort strategy, our method ensures fuel efficiency while maintaining strict safety bounds. The hierarchical, triangulation-based approach identifies topologically distinct, kinematically feasible paths. Validated across controlled and real-world scenarios, the planner adapts to varying current profiles and obstacle configurations, with adaptive padding significantly increasing robustness in high-risk regions.

Future work includes extending this approach to scenarios where the vehicle can adjust its speed, rather than relying on the fixed-effort assumption used in this study. We expect that allowing speed modulation will result in more gliding behavior through the current field. Additional directions include integrating real-time current forecasts or onboard estimation of the vector field, as well as constructing local obstacle maps in partially or fully unknown environments.

\section{Acknowledgments}
This work is supported in part by the Burke Research Initiation Award, NSF CNS-1919647, 2144624, OIA1923004, and NOAA NH Sea Grant.

\bibliography{aaai2026}

\end{document}